\documentclass{article} 
\usepackage{iclr2026_conference,times}

\usepackage{amsmath,amsfonts,bm}

\def\eqref#1{equation~\ref{#1}}

\def\1{\bm{1}}

\DeclareMathAlphabet{\mathsfit}{\encodingdefault}{\sfdefault}{m}{sl}
\SetMathAlphabet{\mathsfit}{bold}{\encodingdefault}{\sfdefault}{bx}{n}

\usepackage{hyperref}
\usepackage{url}
\usepackage{booktabs}
\usepackage{graphicx}
\usepackage{multirow}
\usepackage{array}
\usepackage{placeins}
\usepackage[most]{tcolorbox}
\usepackage{lipsum}
\usepackage{geometry}
\usepackage{float}
\usepackage{caption}
\usepackage{pifont}
\usepackage{xcolor}
\newcommand{\cmark}{\textcolor{green!55!black}{\ding{51}}}
\newcommand{\xmark}{\textcolor{red!75!black}{\ding{55}}}

\title{\textsc{VDiff-Bench}: A Challenging Benchmark for Fine-Grained Image Difference Identification}

\author{Yixin Wan, Tianle Zheng, Kai-Wei Chang \\
Department of Computer Science\\
University of California, Los Angeles\\
\texttt{\{elaine1wan,kwchang\}@cs.ucla.edu} \\
}

\iclrfinalcopy 
\begin{document}
\vspace{-1em}
\maketitle
\begin{figure}[h]
\centering
\vspace{-2em}
\includegraphics[width=1.0\linewidth]{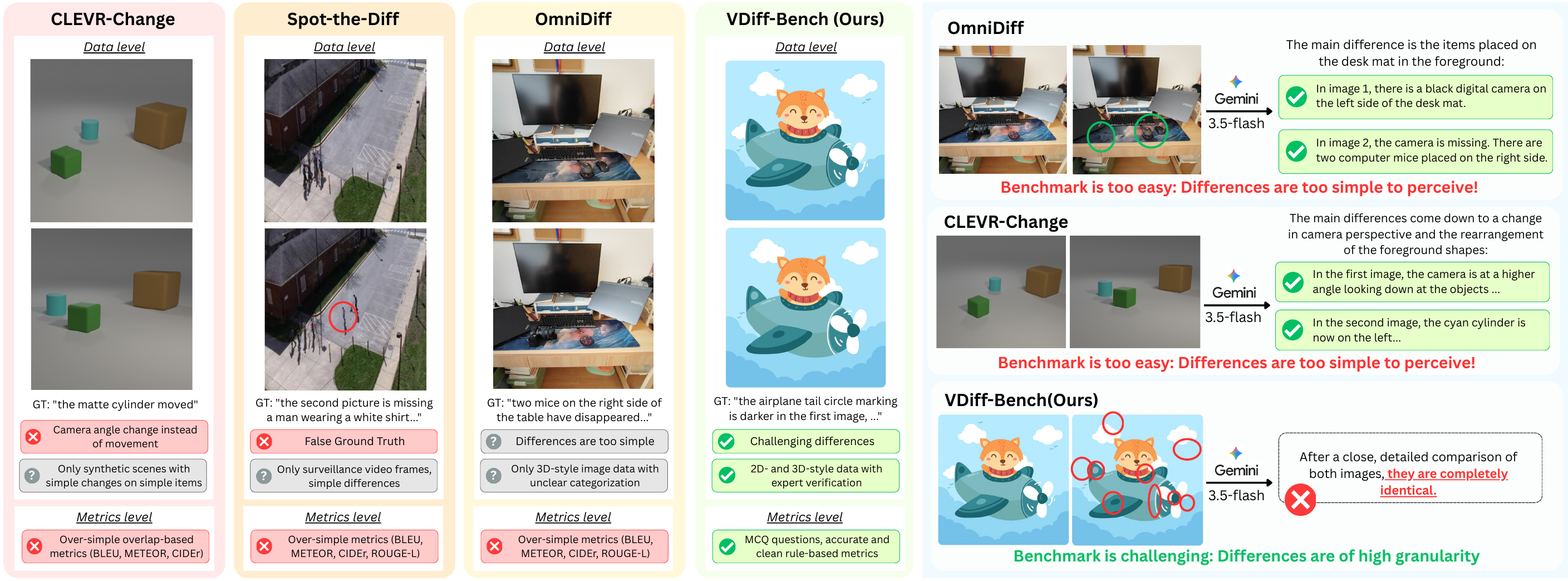}
\caption{
  Comparison with prior visual difference identification benchmarks. Existing benchmarks primarily focus on open-ended difference captioning and are often limited in visual domains, change categories, or evaluation reliability. \textsc{VDiff-Bench} expands to 10 fine-grained change categories and evaluates models through challenging multiple-choice questions for more deterministic and objective IDI performance quantization.
  }
\label{fig:benchmark-comparison}
\end{figure}

\begin{abstract}
Multimodal Large Language Models (MLLMs) perform strongly on general visual understanding tasks such as visual question answering, yet they often struggle with a basic comparative skill: identifying what has changed between two similar images.
We introduce \textbf{\textsc{VDiff-Bench}}\footnote{We release our benchmark at \url{https://huggingface.co/datasets/elaine1wan/image_diff_data}.}, a challenging multiple-choice benchmark for fine-grained Image Difference Identification.
\textsc{VDiff-Bench} contains 1,756 four-way questions over image pairs and covers 10 change categories: position, motion, regional image color, overall image color, appearance/disappearance, noise/resolution, texture, substitution/size, OCR/text, and illumination.
Each question corresponds to two image inputs with 4 choices: the true difference, two hard negative descriptions, and a ``no difference'' distractor.
To make the task challenging, we specifically curate ground-truth-conditioned negatives that require models to distinguish the actual change from nearby semantic alternatives.
Experiments with 11 state-of-the-art open- and closed-source MLLMs show that fine-grained visual comparison remains brittle: models exhibit uneven performance across sources and change categories, with persistent failures on subtle low-level changes like noises and textures.
For instance, 3 7-8B-scale open-source MLLMs score 52.5--70.6\% on semantic changes but only 8.7--33.3\% on low-level changes like noise and texture, falsely assuming no changes between two image inputs.
Surprisingly, despite strong performance of other closed-source commercial models, Grok 4.3 demonstrate remarkable performance drop on identifying noise and texture differences between images, falling significantly behind large open-source models like Kimi K2.5 and K3.
Overall, \textsc{VDiff-Bench} provides a targeted diagnostic for evaluating comparative visual understanding in MLLMs, exposing failures that are not captured by standard single-image vision-language tasks.
Project Page: \url{https://huggingface.co/spaces/elaine1wan/vdiff-bench}.
\end{abstract}

\section{Introduction}
\label{sec:intro}
Multimodal Large Language models (MLLMs) have advanced rapidly on image understanding, achieving strong results on single-image captioning, and visual question answering~\citep{liu2023visualinstructiontuning, bai2023qwenvlversatilevisionlanguagemodel, yue2024mmmu}. 
However, we found that even the strongest closed-source MLLMs fail on the simple task of identifying differences between two similar images---as shown in Figure \ref{fig:benchmark-comparison}---which commonly exist in children's playbooks. 
We refer to this capability as fine-grained \textbf{Image Difference Identification (IDI)}, which requires comparative perception across two views, sensitivity to small localized or low-level differences, and restraint against hallucinating absent changes.

Current benchmarks fail to holistically and accurately assess this capability. 
Existing visual difference benchmarks~\citep{park2019robustchangecaptioning,jhamtani2018learning,liu2025omnidiff} mainly suffer from 2 weaknesses: (1) lack of high-quality, challenging image difference data, and (2) lack of accurate, robust evaluation metrics.
For instance, \citet{park2019robustchangecaptioning} collects scenes synthesized by an image generation engine, but the scenes, objects, and changes are simple and lack diversity.
As shown in the second row of the rightmost examples in Figure \ref{fig:benchmark-comparison}, state-of-the-art MLLMs like Google's Gemini 3.5~\cite{kavukcuoglu2026gemini35} can easily verbalize all differences in these images correctly, even listing out detailed camera angle changes that the benchmark's original ground truth failed to cover.
This suggests that existing benchmarks may no longer be sufficiently challenging to probe the limits of modern MLLMs
Nevertheless, stronger and more holistic visual difference identification benchmarks are crucial for improving MLLMs to support fine-grained perception, image and video editing evaluation, and accurate reward modeling for these generative systems. 
In particular, a strong IDI / IDC model could provide a more grounded signal for tasks like image editing, by comparing visual inputs and identifying what changed, what stayed fixed, and whether the observed changes match the intended transformation.


\begin{figure}[t]
\centering
\vspace{-0.5em}
\includegraphics[width=0.85\linewidth]{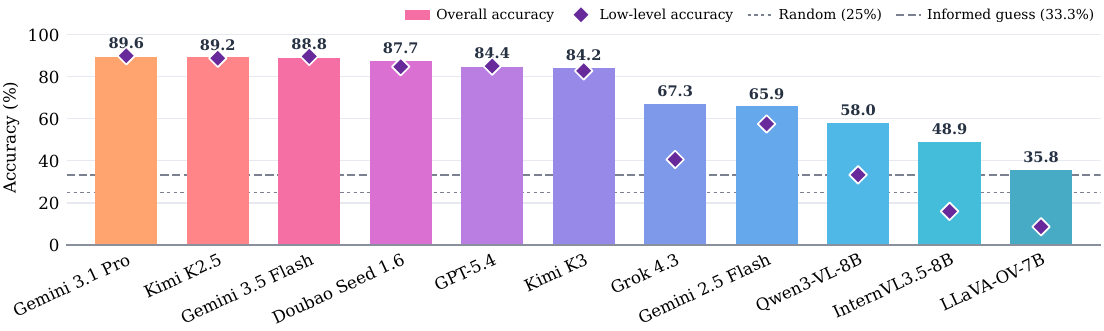}
\vspace{-0.5em}
\caption{\label{fig:headline-results}\textbf{Overall and low-level accuracy on \textsc{VDiff-Bench} across 11 MLLMs.} Bars report overall four-choice accuracy; purple diamonds report accuracy on whole-image color, noise/resolution, texture, and illumination changes. Models are sorted by overall accuracy, and horizontal lines mark the 25\% uniform-guess and 33.3\% informed-guess baselines. Similar overall scores can conceal substantial low-level deficits.}
\vspace{-0.5em}
\end{figure}

\begin{figure*}[t]
  \centering
  \vspace{-0.5em}
  \setlength{\fboxsep}{1.5pt}
  \begin{minipage}[t]{0.49\textwidth}
    \centering
    \textbf{Fine-grained appearance/disappearance}\\[2pt]
    \fbox{\includegraphics[width=0.3\linewidth]{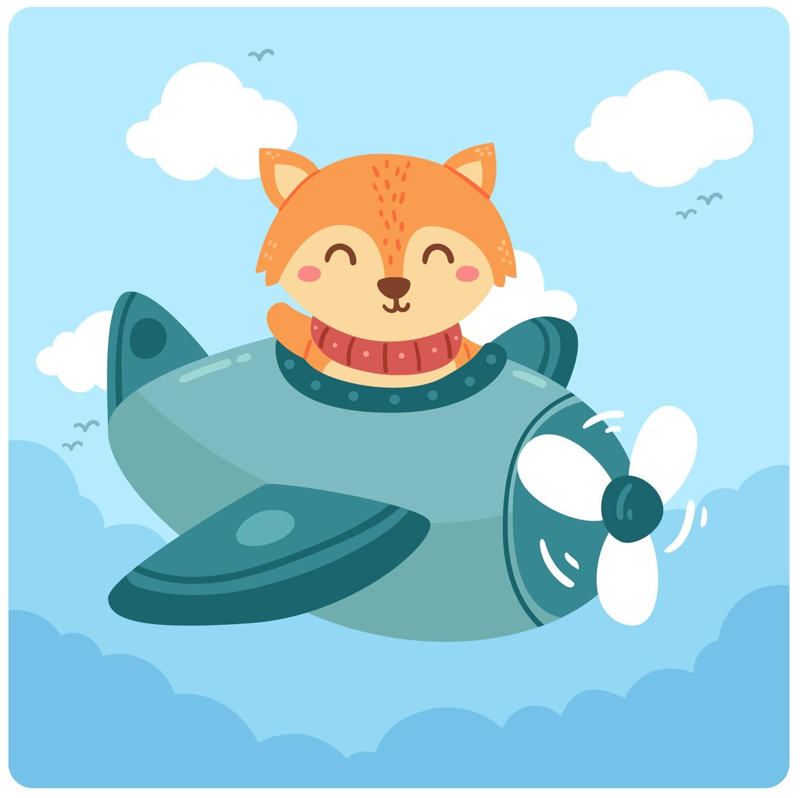}}
    \fbox{\includegraphics[width=0.3\linewidth]{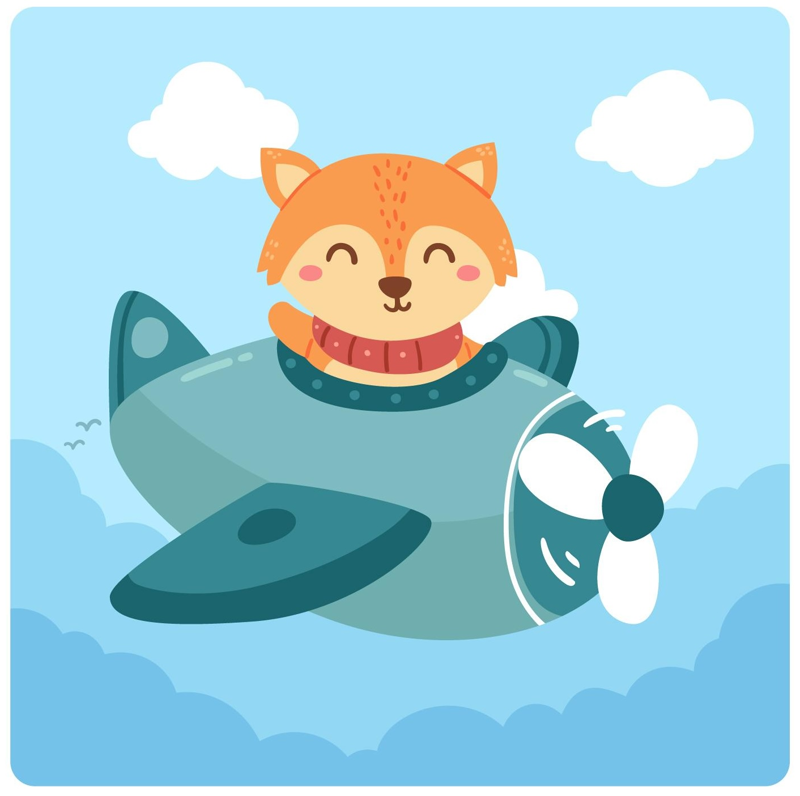}}\\[3pt]
    \raggedright\scriptsize
    \textbf{A. A small bird near the top center is missing in image B.}\\
    B. The birds left of the plane are missing in image B.\\
    C. There is no difference between the two images.\\
    D. The oval wing marking becomes lighter in image B.\\[2pt]
    \textit{Model responses:} GPT-5.4 $\rightarrow$ A; Gemini $\rightarrow$ C; Qwen, InternVL, LLaVA $\rightarrow$ B.
  \end{minipage}\hfill
  \begin{minipage}[t]{0.49\textwidth}
    \centering
    \textbf{Low-level visual texture change}\\[2pt]
    \fbox{\includegraphics[width=0.3\linewidth]{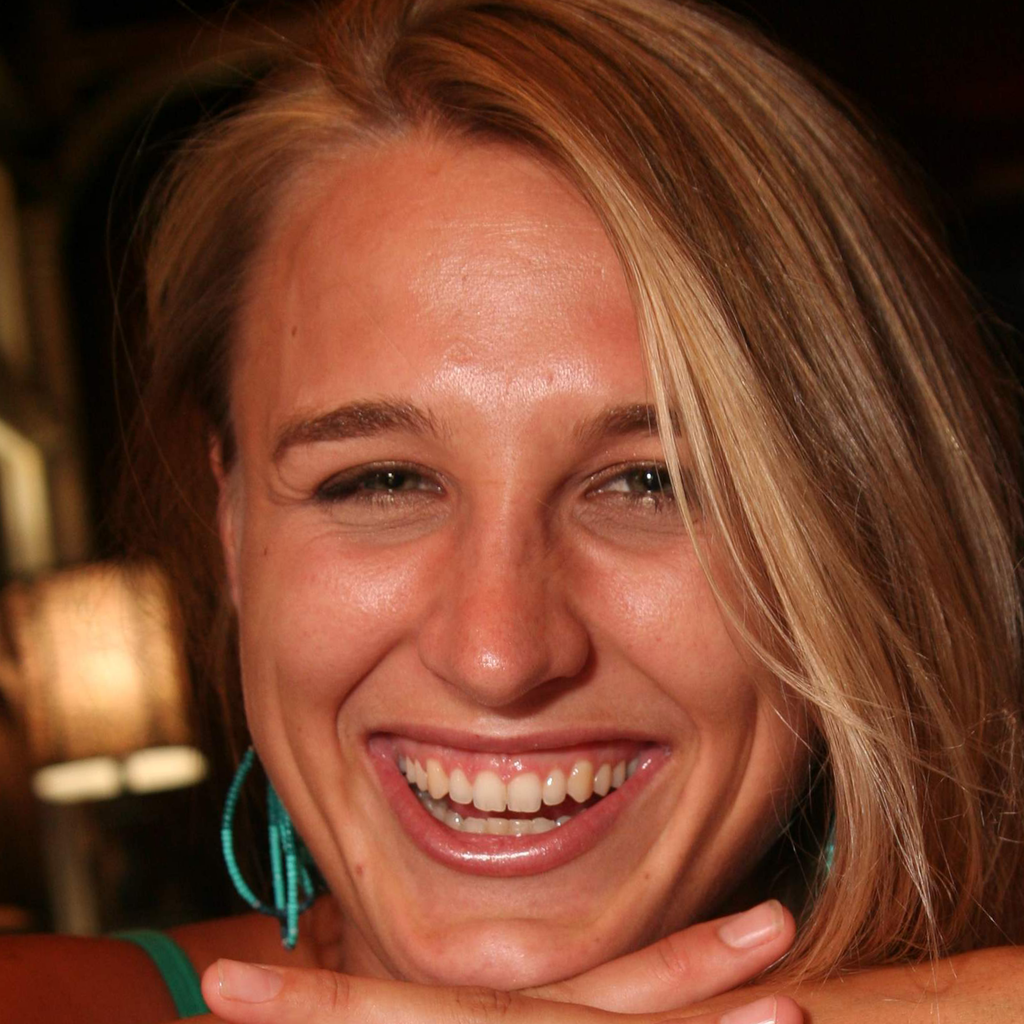}}
    \fbox{\includegraphics[width=0.3\linewidth]{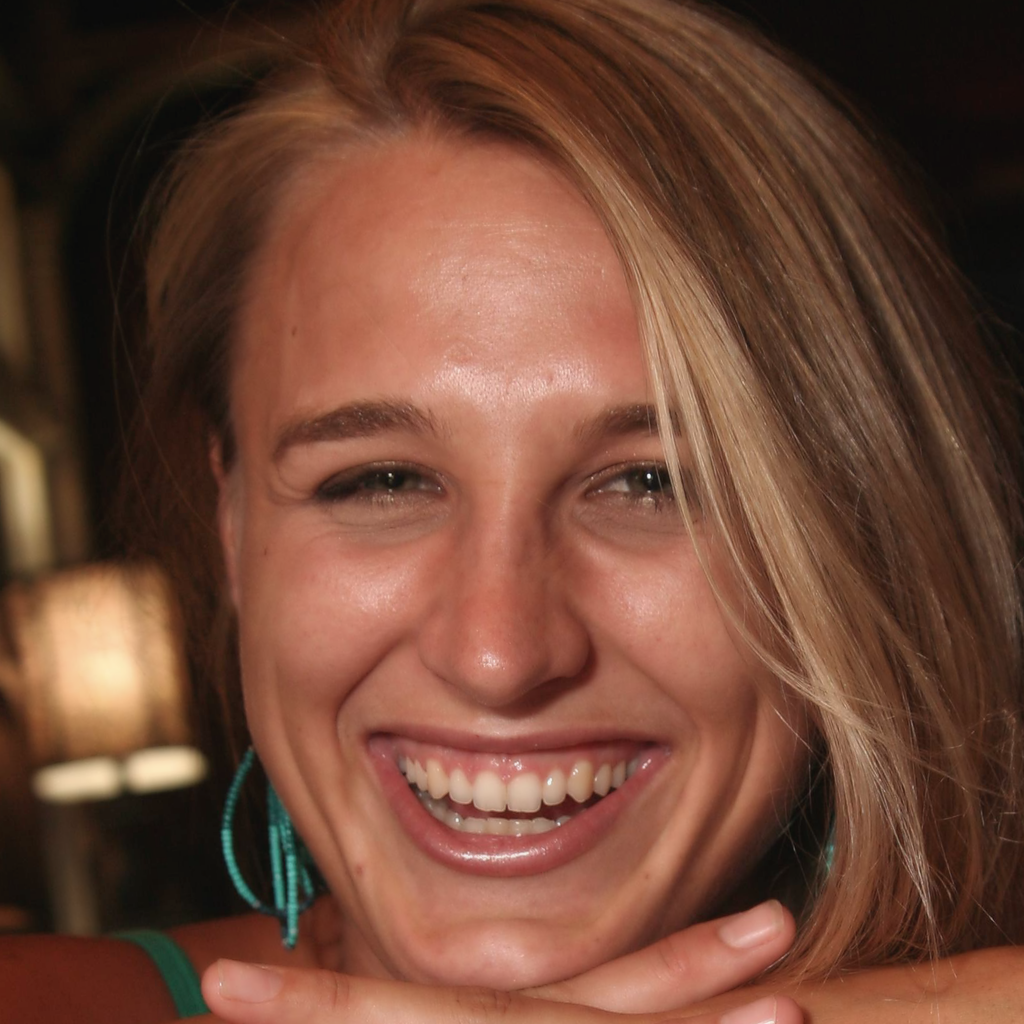}}\\[3pt]
    \raggedright\scriptsize
    \textbf{A. Image B is smoother than image A.}\\
    B. The woman's hair changes from blonde to light brown.\\
    C. There is no difference between the two images.\\
    D. Image B is sharper and more detailed than image A.\\[2pt]
    \textit{Model responses:} GPT-5.4 $\rightarrow$ A; Gemini, Qwen, InternVL, LLaVA $\rightarrow$ C.
  \end{minipage}
  \caption{Each evaluation example in \textsc{VDiff-Bench} contains a ground truth difference, two false difference distractor options, and one no-difference distractor option. We show an example of higher-level direct visual change (i.e. changing an object clearly visible) on the left and an example of lower-level visual change (i.e. changing the image's trait like texture) on the right.}
  \label{fig:VDiff-Bench-overview}
  \vspace{-0.7em}
\end{figure*}

To address the research gap on challenging IDI benchmarks, we introduce \textbf{\textsc{VDiff-Bench}}, a diagnostic multiple-choice benchmark designed for this purpose.
It contains 1,756 questions over distinct image pairs, organized into 10 change categories: position, motion, regional image color, whole-image color, appearance/disappearance, noise/resolution, texture, substitution/size, OCR/text, and illumination.
Rather than scoring free-form captions, we reformulate the task as a multiple-choice question (MCQ): each item presents an aligned image pair and asks the model to select the single option that states a real difference, against an explicit ``no difference'' option and several plausible-but-false distractor choices constructed with extensive human verification and correction.
For instance, Figure \ref{fig:VDiff-Bench-overview} shows two examples in the \textsc{VDiff-Bench} benchmark with challenging distractor options.

Evaluation across 11 contemporary MLLMs reveals pronounced category-specific brittleness that does not follow a simple proprietary-versus-open-weight divide. Three open 7--8B models achieve 52.5--70.6\% accuracy on semantic changes but only 8.7--33.3\% on low-level changes, selecting the ``no difference'' distractor on 51.3--80.9\% of low-level questions.
Specifically, \textbf{these models falsely select the ``no difference'' distractor choice on 51.3--80.9\% of low-level questions}, showing major limitation in difference perception capabilities, suggesting that model capacity remains an important bottleneck.
Yet scale alone is insufficient: while the larger-scale Kimi K2.5 and Kimi K3 attain 88.8\% and 82.8\% low-level visual difference accuracy, respectively, Grok 4.3, a closed-source large-scale commercial model, achieves only 40.7\%--only 5.3\% accuracy on noise difference category and only 15.3\% accuracy on texture difference category.
These results are consistent with a two-factor account: scale may raise the attainable ceiling, but training data, learning objectives, and visual encoding might determine whether that capacity translates into precise cross-image comparison. Fine-grained visual difference identification therefore appears not to be an automatic consequence of general multimodal scaling, but a distinct capability that must be explicitly developed and evaluated during training

Our contributions are threefold:
\begin{itemize}
  \item We introduce \textsc{VDiff-Bench}, a 1,756-question Image Difference Identification (IDI) benchmark spanning ten categories across semantic, textual, and low-level changes.
  \item We formulate IDC into IDI, a multiple choice task with a ground truth answer and three distractor choices, enabling deterministic scoring without a questionable response-level judge proposed by prior works.
  \item We systematically evaluate 11 MLLMs with category-, pair-, and response-level analyses, revealing descriptive semantic--low-level gaps.
\end{itemize}

\section{Related Work}
\label{sec:related}
\subsection{General MLLM Benchmarks}
Evaluation of MLLMs' general capabilities has largely been conducted around single-image descriptive tasks such as Visual Question-Answering (VQA), Visual Reasoning, etc..
For instance, VQA established the task of answering natural-language questions about images~\cite{agrawal2016vqavisualquestionanswering}, with later benchmarks extending it to compositional and relational reasoning~\cite{hudson2019gqanewdatasetrealworld}.
On the reasoning side, previous works have evaluated MLLM's ability to reason on mathematical and diagrammatic reasoning~\cite{lu2024mathvistaevaluatingmathematicalreasoning,zhang2024mathverse,wang2024measuring}, as well as broad college-level multimodal knowledge~\cite{yue2024mmmu}.
These works motivate evaluating MLLMs beyond high-level semantic recognition, especially on tasks requiring subtle visual comparison.


\subsection{Image Difference Identification and Captioning Benchmarks}
A series of works extend MLLM evaluation to multi-image, comparative scenarios.
Specifically, the task of Image difference captioning (IDC) prompts a model to identify changes between paired images.
For instance, Spot-the-Diff~\citep{jhamtani2018learning} introduced image pairs from surveillance footages with crowd-sourced image descriptions.
CLEVR-Change~\citep{park2019robustchangecaptioning} utilized rendered images from synthetic scenes with five object-change types.
These datasets test MLLMs on change localization and verbalization, but their domains and change inventories are limited.
More recently, OmniDiff~\citep{liu2025omnidiff} broadens IDC to real and rendered image pairs across scenarios and change types, with human descriptions.
However, it evaluates model-verbalized differences using inaccurate reference-based metrics like BLEU-4 and ROUGE-l, which remain sensitive to paraphrase and do not cleanly attribute omitted, reversed, or unsupported claims.
DiffCap-Bench~\citep{wei2026diffcapbench} addresses this issue by using MLLM judge-reported metrics~\citep{wei2026diffcapbench}, but this method relies heavily on the performance of the LLM judge---while a large body of previous works~\citep{NEURIPS2023_91f18a12,wang-etal-2024-large-language-models-fair,NEURIPS2024_7f1f0218,raina-etal-2024-llm} have revealed significant issues with lack of robustness and biases in LLM judges.
\textsc{VDiff-Bench} is complementary to both: it does not assess free-form completeness, but converts one selected change into a controlled discrimination problem with deterministic scoring.


\subsection{MLLMs for Image Editing Evaluation}
Automatic evaluation of image editing models has always been a difficult yet important task.
Early works~\citep{xu2023imagereward,Kirstain2023PickaPicAO} learn a preference score model from human feedback.
However, as the generation scene become increasingly compositional and complex, recent image editing models have widely adopted MLLM judges for evaluating generated image quality~\citep{ye2025imgedit,li2025uniworldv2}. 
For image editing evaluation, fine-grained visual comparison is crucial: an evaluator must verify that the requested modification occurred while detecting incorrect, unintended changes (e.g. background). 
However, MLLM judges constantly fails to accurately describe fine-grained edit-induced visual differences, frequently hallucinating changes~\citep{yosef-etal-2025-editinspector}.
This motivates for dedicated benchmarks and methods for the IDC task.

\section{The \textsc{VDiff-Bench} Benchmark}
\label{sec:benchmark}
The construction of \textsc{VDiff-Bench} proceeds in three stages:
First, we collect image pairs from existing datasets, manually annotate un-labeled image pair data, as well as create synthetic images to augment under-represented change categories.
Second, we synthesize false differences between images as distractor option candidates, and conduct human verification, filtering, and re-writing for low-quality ones.
Third, we aggregate the ground truth difference description with the sampled distractor options to construct the multiple choice-format evaluation data in \textsc{VDiff-Bench}.
Below, we elaborate on our task definition, data sources, and the data construction process.

\subsection{Task Definition}
\label{sec:task}

Each \textsc{VDiff-Bench} data entry consists of an ordered image pair $(I_A,I_B)$ and four textual options $\mathcal{O}=\{o_A,o_B,o_C,o_D\}$.
The benchmark assigns one option as a reference-supported difference, two as candidate alternatives intended to be false, and one fixed distractor option claiming that the two images are completelhy identical.
All \textsc{VDiff-Bench} pairs contain at least one real change, so the no-difference option is always a distractor.
The model returns a label $\hat{y}\in\{A,B,C,D\}$, and we report choice accuracy:
\begin{equation}
  \operatorname{Acc}=\frac{1}{N}\sum_{i=1}^{N}\mathbb{1}[\hat{y}_i=y_i].
\end{equation}
Among parsed incorrect choices, selecting no difference records a \emph{missed-change selection}, whereas selecting either candidate alternative records a competing-change selection.

\subsection{Image-Pair Collection and Taxonomy}
\label{sec:data}
\textsc{VDiff-Bench} collects image-difference pairs organized into 10 difference categories, where each subset denotes a distinct change condition.
The 10 categories cover both semantic edits and changes that depend more strongly on low-level comparative vision perception.

\begin{table}[h]
  \centering
  \scriptsize
  \setlength{\tabcolsep}{5pt}
  \begin{tabular}{lrl}
    \toprule
    \textbf{Category} & \textbf{\# Q.} & \textbf{Representative distinction} \\
    \midrule
    Position & 158 & left/right, nearer/farther, relocated object \\
    Motion & 138 & action, pose, orientation, or state transition \\
    Regional color & 170 & localized object or region color \\
    Whole-image color & 147 & global RGB or color-cast shift \\
    Appearance/disappearance & 201 & added, removed, or missing entity \\
    Noise/resolution & 150 & Gaussian noise or resolution degradation \\
    Texture & 177 & smoothing, pattern, mark, or surface detail \\
    Substitution/size & 199 & object replacement, count, or scale \\
    OCR/text & 221 & character-, digit-, or word-level change \\
    Illumination & 195 & global or scene-level brightness change \\
    \midrule
    \textbf{Total} & \textbf{1,756} & \\
    \bottomrule
  \end{tabular}
  \caption{Current \textsc{VDiff-Bench} taxonomy. Counts refer to questions; the present inventory contains 1,756 questions over 1,543 distinct image pairs.}
  \label{tab:categories}
  \vspace{-0.8em}
\end{table}

\subsubsection{Data Sources}
Our raw image data consists of existing annotated and un-annotated pairs, as well as unpaired image data for change augmentation.
Appendix~\ref{app:sources} reports the source composition and additional construction details.
\vspace{-0.8em}

\paragraph{Existing annotated pairs.}
We draw from complementary paired-image resources: fixed-camera scenes from Spot-the-Diff~\citep{jhamtani2018learning}, motion-centric edits from MotionEdit~\citep{wan2025motionedit}, color and position edits from OmniEdit~\citep{wei2025omniedit}, and illumination, substitution, size, and OCR cases from OmniDiff~\citep{liu2025omnidiff}.
We use source difference annotations or editing instructions as provenance for the intended change, then normalize each selected statement and map it to the current \textsc{VDiff-Bench} taxonomy.
\vspace{-0.8em}

\paragraph{Existing un-annotated pairs.}
Inspired by the ``spot-the-difference'' puzzle games in childrens' playbooks, we collect a set of 196 ``find-the-difference'' puzzle pairs curated from various online sources. 
These drawn cartoon scenes contain many small and deliberately challenging visual differences per pair of images.
\vspace{-0.8em}

\paragraph{Unpaird image data.}
Inspired by the user-identified failure modes in image editing models to preserve human skin texture~\citep{smith2026aieditedface}, we sample from the FFHQ dataset~\citep{karras2019style} with high-quality facial images and augment them by applying low-level visual changes (see \ref{sec:data-augmentation}).
Additionally, we augment our dataset with state-of-the-art image editing models on scene-text images from MLT19~\citep{nayef2019icdarmlt} and TextOCR~\citep{singh2021textocr}, as well as sampled input images from OmniEdit~\citep{wei2025omniedit}.

\subsubsection{Data Augmentation}
\label{sec:data-augmentation}
\paragraph{Programmatic Transformations}
We augment our evaluation benchmark by applying programmatic transformations on low-level vision features for FFHQ facial images.
These include applying sampled Gaussian noise perturbations, smoothing texture changes, whole-image RGB shifts, and gamma/linear illumination changes.
Because the transformation is applied programmatically, its transformation type and direction natually provides the ground truth for difference captioning.
\vspace{-0.8em}

\paragraph{Image Editing}
Additionally, we augment the under-represented position difference and OCR/text difference image data by conducting image editing on source images.
For position difference, we sequentially sampled source images from OmniEdit’s object-swap subset, generated new image-conditioned position-edit instructions with GPT-5.4-mini, and applied the first proposed instruction using GPT-image-2~\citep{openai2026gptimage2}.
For OCR/text differences, we manually curate localized text-edit instructions for sampled scene-text images and apply them using Gemini-3-Pro-Image~\citep{geminiteam2026gemini31pro}.
Image difference ground truth for these data are directly derived from the editing prompts.
\vspace{-0.8em}

\paragraph{Human Annotation and Cross-Validation}
\label{sec:annotation}
For existing un-annotated pairs curated puzzle pairs without sufficiently detailed source annotations, we collected ground-truth difference descriptions from volunteer domain experts who are fluent in English. 
Annotators inspected each ordered image pair side by side and enumerated all visible differences. Each description was required to identify a single change, specify the affected object or region using distinguishing visual attributes and spatial cues, and state explicitly how the first and second images differ. 
Annotators were asked to capture not only added or missing objects, but also localized changes in color, shape, orientation, and fine-grained pattern or texture, while avoiding speculative or overly vague descriptions.
To ensure the quality of the annotated image differences, every annotation subsequently underwent a second review to be cross-validated by a different expert. The reviewer re-examined the image pair for coverage and visual support, and rewrote descriptions that were inaccurate, ambiguous, overly broad, grammatically unclear, or inconsistent in comparison direction. The resulting reviewed descriptions form the ground-truth pool from which reference differences are selected during question construction.

\begin{table*}[t]
  \centering
  \small
  \setlength{\tabcolsep}{3.5pt}
  \resizebox{\textwidth}{!}{%
  \begin{tabular}{l r cccc ccccc cc}
    \toprule
    \multirow{3}{*}{\textbf{Benchmark}}
    & \multirow{3}{*}{\textbf{Pairs}}
    & \multicolumn{4}{c}{\textbf{Image Regime}}
    & \multicolumn{5}{c}{\textbf{Target-Change Coverage}}
    & \multicolumn{2}{c}{\textbf{Evaluation Design}} \\
    \cmidrule(lr){3-6}
    \cmidrule(lr){7-11}
    \cmidrule(lr){12-13}
    &
    & \textbf{Real}
    & \textbf{Edited}
    & \textbf{Rendered}
    & \shortstack{\textbf{2D}\\\textbf{puzzle}}
    & \textbf{Semantic}
    & \shortstack{\textbf{OCR}/\\\textbf{text}}
    & \shortstack{\textbf{Global}\\\textbf{photo.}}
    & \shortstack{\textbf{Noise}/\\\textbf{res.}}
    & \textbf{Texture}
    & \shortstack{\textbf{Hard}\\\textbf{alternatives}}
    & \shortstack{\textbf{Exact scoring}} \\
    \midrule
    Spot-the-Diff~\citep{jhamtani2018learning}
      & 13,192 & \cmark & \xmark & \xmark & \xmark
      & \cmark & \xmark & \xmark & \xmark & \xmark
      & \xmark & \xmark \\
    CLEVR-Change~\citep{park2019robustchangecaptioning}
      & 79,606 & \xmark & \xmark & \cmark & \xmark
      & \cmark & \xmark & \xmark & \xmark & \cmark
      & \xmark & \xmark \\
    OmniDiff~\citep{liu2025omnidiff}
      & 15,598 & \cmark & \xmark & \cmark & \xmark
      & \cmark & \cmark & \cmark & \xmark & \xmark
      & \xmark & \xmark \\
    DiffCap-Bench~\citep{wei2026diffcapbench}
      & 1,075 & \cmark & \cmark & \cmark & \xmark
      & \cmark & \cmark & \cmark & \cmark & \cmark
      & \xmark & \xmark \\
    \textbf{VDiff-Bench (ours)}
      & \textbf{1,543} & \cmark & \cmark & \cmark & \cmark
      & \cmark & \cmark & \cmark & \cmark & \cmark
      & \cmark & \cmark \\
    \bottomrule
  \end{tabular}}
  \caption{\textbf{Coverage and evaluation design of visual difference benchmarks.}
  Check marks indicate dimensions explicitly included in benchmark construction or evaluated as target differences.
  ``Global photo.'' includes whole-image color and illumination changes, while ``Noise/res.'' includes noise or resolution degradation.
  Hard alternatives are plausible competing change statements, and exact, judge-free scoring denotes direct answer-key matching without reference-caption metrics or a learned evaluator.}
  \label{tab:benchmark-comparison}
\end{table*}

\subsection{False Difference Generation}
\label{sec:construction}
To challenge MLLMs on the IDI task, we construct false image differences that are semantically plausible as distractor options for models.
\vspace{-0.8em}

\paragraph{Structured False Difference Generation}
We first generate a set of false differences with reference to the real difference annotations in image pairs.
Specifically, we utilize two strong MLLMs--Gemini 2.5 Pro and GPT-5.5--by providing them with image pair inputs and their reference ground truth difference lists, and ask it to generate a list of at least 3 false differences intended to be incorrect but visually plausible.
We instruct the model to anchor each false difference generation on one ground truth difference, using strategies like applying the reference change to a nearby entity, reversing a direction or state, or substituting a plausible attribute while preserving the scene vocabulary.
This procedure is designed to produce candidate alternatives close to the reference in content and phrasing, rather than unrelated answer options.
Finally, structured filters remove duplicates and exact truth matches.
\vspace{-0.8em}

\paragraph{Human Verification}
Model-generated false differences might be too semantically unplausible or too vague to be judged, therefore not acting as challenging ``negative'' choices that an evaluated MLLM needs to distinguish from.
Therefore, we invite a human expert to inspect the image pairs and generated false differences and refine, rewrite, or discard low-quality ones.

\subsection{Multiple-Choice Question Construction}
To construct the final multiple-choice questions in our \textsc{VDiff-Bench} dataset, we retain the ground truth difference caption, two false difference statements for each image pair, append the fixed no-difference distractor option, and shuffle the four options to be randomly ordered.
To control for option-position bias and rule out fixed response strategies (e.g., always selecting option D), we shuffle the four choices using a fixed random seed; consequently, both the correct-answer labels and the no-difference distractor positions are approximately balanced across A--D.
Appendix~\ref{app:construction} provides the prompt and additional audit statistics.


\subsection{Dataset Statistics}
Our final \textsc{VDiff-Bench} benchmark consists of \textbf{1,756} questions across \textbf{10 image difference categories}.
Table~\ref{tab:categories} defines the 10 categories and gives their question counts.
The image differences span both \textit{semantic changes}--- position, motion, regional color, appearance/disappearance, substitution/size, and OCR/text changes---as well as \textit{low-level visual trait changes} like whole-image color, noise/resolution, texture, and illumination.

\vspace{-0.8em}
\subsection{Comparison with Existing Benchmarks}
\label{sec:comparison}
We compare \textsc{VDiff-Bench} against four direct image-difference-captioning benchmarks in Table~\ref{tab:benchmark-comparison}. As the table shows, existing benchmarks provide valuable scale and diversity but leave 2 major gaps. First, none jointly evaluates semantic, textual, and low-level target changes across real, edited, rendered, and densely composed 2D puzzle images. Second, they formulate evaluation as free-form caption generation, requiring either reference-caption metrics or an MLLM judge to determine whether a predicted difference is correct. \textsc{VDiff-Bench} addresses the coverage gap by bringing these image regimes and change families into a unified taxonomy, and addresses the evaluation gap by introducing human-verified, reference-conditioned alternatives with exact, judge-free choice scoring. This formulation directly tests whether a model can distinguish the observed change from plausible but unsupported alternatives, complementing prior benchmarks that measure the completeness and quality of free-form descriptions.


\begin{table}[t]
  \centering
  \scriptsize
  \setlength{\tabcolsep}{3.2pt}
  \resizebox{\textwidth}{!}{%
  \begin{tabular}{lccccccccccc}
    \toprule
    \midrule
    \multirow{2}{*}{\textbf{Model}} & \multicolumn{10}{c}{\textbf{Change Category}} & \multirow{2}{*}{\textbf{Overall}} \\
    \cmidrule{2-11}
     & \textbf{Pos.} & \textbf{Motion} & \textbf{Reg. col.} & \textbf{Whole col.} & \textbf{App./dis.} & \textbf{Noise} & \textbf{Texture} & \textbf{Sub./size} & \textbf{OCR} & \textbf{Illum.} &  \\
    \midrule
    \multicolumn{12}{c}{\textit{Baselines}} \\
    \midrule
    Random & 25.0 & 25.0 & 25.0 & 25.0 & 25.0 & 25.0 & 25.0 & 25.0 & 25.0 & 25.0 & 25.0 \\
    Informed guess & 33.3 & 33.3 & 33.3 & 33.3 & 33.3 & 33.3 & 33.3 & 33.3 & 33.3 & 33.3 & 33.3 \\
    Text-only style & 57.6 & 33.3 & 34.1 & 27.2 & 61.2 & 32.7 & 32.8 & 35.7 & 33.5 & 31.3 & 38.2 \\
    \midrule
    \multicolumn{12}{c}{\textit{Closed Source Models}} \\
    \midrule
    GPT-5.4 & 83.5 & 92.8 & 92.4 & 83.7 & 61.7 & \textbf{100.0} & 83.1 & 91.5 & 86.0 & 76.4 & 84.4 \\
    Gemini 2.5 Flash & 57.0 & 81.9 & 82.9 & 57.8 & 44.8 & 89.3 & 26.6 & 74.9 & 86.0 & 61.0 & 65.9 \\
    Gemini 3.1 Pro & \textbf{88.6} & 91.3 & 94.7 & 86.4 & 75.1 & \textbf{100.0} & \textbf{87.6} & 94.5 & \textbf{92.8} & 87.2 & \textbf{89.6} \\
    Gemini 3.5 Flash & 88.0 & 88.4 & 92.9 & 85.7 & \textbf{79.6} & \textbf{100.0} & 83.1 & 92.5 & 89.1 & \textbf{90.8} & 88.8 \\
    Grok 4.3 & 75.9 & 96.4 & 89.4 & 87.1 & 68.7 & 5.3 & 15.3 & 83.9 & 90.0 & 55.9 & 67.3 \\
    Doubao Seed 1.6 Vision & 86.1 & 95.7 & 94.7 & 91.8 & 72.6 & 97.3 & 76.8 & \textbf{97.5} & 92.3 & 76.9 & 87.7 \\
    \midrule
    \multicolumn{12}{c}{\textit{Open Source Models}} \\
    \midrule
    Qwen3-VL-8B Thinking & 59.5 & 87.7 & 86.5 & 21.8 & 53.2 & 45.3 & 8.5 & 71.4 & 83.3 & 55.4 & 58.0 \\
    InternVL3.5-8B & 56.3 & 84.8 & 85.3 & 6.1 & 52.7 & 3.3 & 3.4 & 73.9 & 66.5 & 44.6 & 48.9 \\
    LLaVA-OneVision-Qwen2-7B & 43.7 & 74.6 & 73.5 & 1.4 & 32.8 & 7.3 & 5.1 & 46.2 & 52.5 & 18.5 & 35.8 \\
    Kimi K2.5 & 84.2 & 94.9 & \textbf{95.9} & \textbf{93.9} & 73.6 & \textbf{100.0} & 83.1 & \textbf{97.5} & 92.3 & 81.5 & 89.2 \\
    Kimi K3 & 80.4 & \textbf{97.1} & 93.5 & 82.3 & 64.2 & \textbf{100.0} & 78.5 & 89.9 & 89.1 & 73.8 & 84.2 \\
    \midrule
    \bottomrule
  \end{tabular}}
  \caption{\textbf{Choice accuracy (\%) on \textsc{VDiff-Bench}.} Results are reported by change category and overall; invalid outputs count as incorrect. Random guessing over four options yields 25.0\%, while informed guessing uniformly selects among the three descriptive options, exploiting that ``no difference'' is always incorrect, and yields 33.3\%. The text-only baseline uses only capitalization and punctuation cues without viewing the images. Bold indicates the best MLLM performance in each column, including ties.}
  \label{tab:category-results}
  \vspace{-0.8em}
\end{table}
\section{Experiments}
\label{sec:experiments}

\subsection{Experimental Setup}
\label{sec:setup}

We evaluate six proprietary MLLMs: GPT-5.4 (snapshot
\texttt{2026-03-05})~\citep{openai2026gpt54}, Gemini 2.5
Flash~\citep{comanici2025gemini25}, Gemini 3.1 Pro
(Preview)~\citep{geminiteam2026gemini31pro}, Gemini 3.5
Flash~\citep{kavukcuoglu2026gemini35}, Grok
4.3~\citep{xai2026grok43}, and Doubao Seed 1.6
Vision~\citep{bytedanceseed2025seed16}.
For Doubao Seed 1.6 Vision, we use the
\texttt{doubao-seed-1-6-vision-250815} snapshot.
We additionally evaluate five open-weight MLLMs: Qwen3-VL-8B
Thinking~\citep{bai2025qwen3vl}, InternVL3.5-8B~\citep{wang2025internvl35},
LLaVA-OneVision-Qwen2-7B~\citep{li2024llavaonevision}, Kimi
K2.5~\citep{kimiteam2026kimik25}, and Kimi
K3~\citep{kimiteam2026kimik3}.
\vspace{-0.8em}
\paragraph{Generation Setup.}
Every model receives images A and B in that order together with four labeled options and the instruction to return one label only.
For open-source models, we set the generation temperature to 0
Appendix~\ref{app:prompt} provides details on the prompt template and run configuration.
\vspace{-0.8em}

\paragraph{Evaluation Metrics.}
We report the answer-key choice accuracy as our main metric. 
To conduct stratified analysis, we report the overall accuracy, accuracy by change category, as well as aggregated accuracy for semantic difference groups and low-level visual difference groups as defined in Section~\ref{sec:data}.

\begin{figure}[h]
  \centering
  \includegraphics[width=0.75\textwidth]{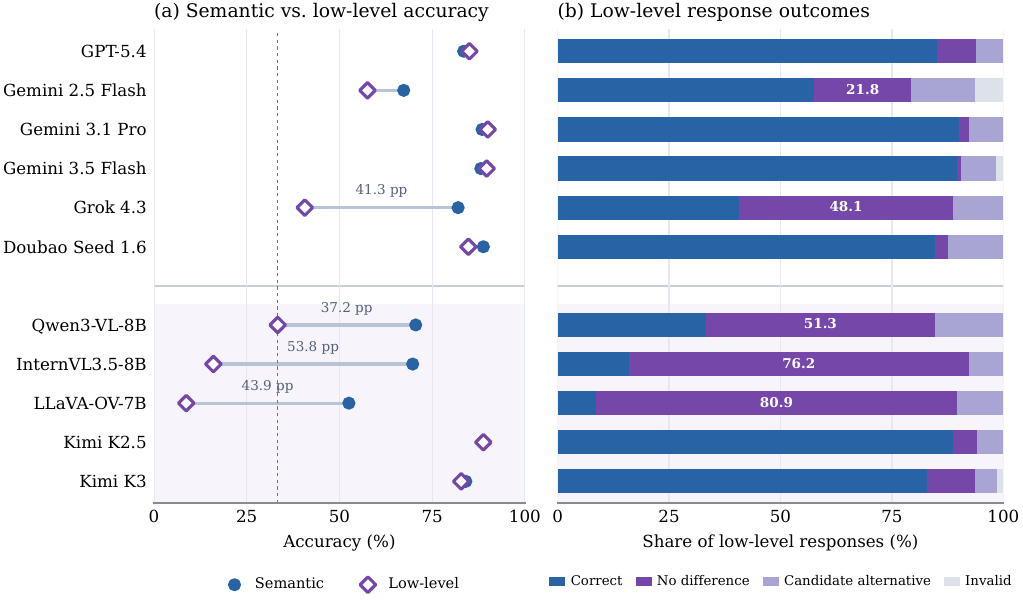}
  \vspace{-0.5em}
  \caption{\label{fig:group-performance}Semantic and low-level performance across 11 MLLMs. (a) Accuracy on semantic and low-level change groups; connectors expose within-model gaps, labels mark gaps of at least 25 percentage points, and the dashed line denotes the 33.3\% informed-guess baseline. (b) Low-level predictions decomposed into correct, no-difference, candidate-alternative, and invalid outcomes; labels mark no-difference rates of at least 15\%.}
  \vspace{-1em}
\end{figure}

\subsection{Results by Change Category}
\label{sec:results}
Table~\ref{tab:category-results} presents evaluation results by different change categories, as well as the overall result.
\vspace{-0.8em}

\paragraph{Models diverge substantially in overall VDI performance.}
Overall accuracy ranges from 35.8\% to 89.6\%, compared with the 25\% uniform-guess and 33.3\% informed-guess baselines.
Gemini 3.1 Pro ranks first at 89.6\% (pair-clustered 95\% CI: 87.9--91.2), narrowly followed by Kimi K2.5 at 89.2\% (87.4--91.0), Gemini 3.5 Flash at 88.8\% (87.2--90.4), and Doubao Seed 1.6 Vision at 87.7\% (85.7--89.7).
No model dominates the taxonomy: the best category scores are distributed across Gemini, Kimi, and Doubao models, indicating that similar aggregate accuracy can conceal distinct perceptual strengths.
\vspace{-0.8em}

\paragraph{Model scale is enabling but not sufficient for low-level visual comparison.}
The 7--8B open-weight models perform substantially worse on low-level than semantic changes, suggesting a capacity bottleneck at smaller scales.
Yet scale alone does not explain the results: Kimi K2.5 and Kimi K3 achieve 88.8\% and 82.8\% low-level accuracy, respectively, whereas Grok 4.3 falls from 82.0\% on semantic changes to 40.7\% on low-level changes, scoring only 5.3\% on noise difference category and 15.3\% on texture difference category.
This shows that increasing model scale does not guarantee strong VDI capability.
\vspace{-0.8em}

\paragraph{Category rankings reveal failure modes.}
Grok 4.3 reaches 96.4\% on motion but only 5.3\% on noise/resolution and 15.3\% on texture, while Kimi K2.5 leads on regional and whole-image color and ties for the best substitution/size result.
Appearance/disappearance remains difficult even for the strongest systems: the category maximum is 79.6\%, compared with at least 90.8\% in seven other categories.
In Appendix \ref{appendix:additional-qualitative-results}, Figure~\ref{fig:qualitative-atlas}, we provide additional qualitative results of different models on data from each change category.

\begin{table}[h]
  \centering
  \scriptsize
  \setlength{\tabcolsep}{4pt}
  \resizebox{0.75\linewidth}{!}{%
  \begin{tabular}{l|c|ccc|cc|c}
    \toprule
    \midrule
    \multirow{2}{*}{\textbf{Model}} & \multirow{2}{*}{\textbf{Pair macro}} & \multicolumn{3}{c}{\textbf{Change Group}} & \multicolumn{3}{c}{\textbf{Error Type}} \\
    \cmidrule{3-5} \cmidrule{6-8}
     &  & \textbf{Semantic} & \textbf{Low-level} & \textbf{OCR} & \textbf{No diff.} & \textbf{Alternative} & \textbf{Invalid} \\
    \midrule
    \multicolumn{8}{c}{\textit{Closed Source Models}} \\
    \midrule
    GPT-5.4 & 86.2 & 83.5 & 85.1 & 86.0 & 5.2 & 10.4 & 0.0 \\
    Gemini 2.5 Flash & 67.4 & 67.3 & 57.5 & 86.0 & 12.1 & 18.6 & 3.4 \\
    Gemini 3.1 Pro & 91.4 & 88.5 & 90.0 & 92.8 & 1.6 & 8.3 & 0.6 \\
    Gemini 3.5 Flash & 90.4 & 88.1 & 89.7 & 89.1 & 0.5 & 7.8 & 2.9 \\
    Grok 4.3 & 66.2 & 82.0 & 40.7 & 90.0 & 21.8 & 10.9 & 0.0 \\
    Doubao Seed 1.6 Vision & 89.6 & 88.8 & 84.8 & 92.3 & 1.3 & 11.0 & 0.0 \\
    \midrule
    \multicolumn{8}{c}{\textit{Open Source Models}} \\
    \midrule
    Qwen3-VL-8B Thinking & 57.6 & 70.6 & 33.3 & 83.3 & 24.8 & 17.2 & 0.0 \\
    InternVL3.5-8B & 47.3 & 69.7 & 16.0 & 66.5 & 36.8 & 14.3 & 0.0 \\
    LLaVA-OneVision-Qwen2-7B  & 34.5 & 52.5 & 8.7 & 52.5 & 43.1 & 21.1 & 0.0 \\
    Kimi K2.5 & 91.2 & 88.8 & 88.8 & 92.3 & 3.4 & 7.4 & 0.0 \\
    Kimi K3 & 86.2 & 84.1 & 82.8 & 89.1 & 5.9 & 4.7 & 5.2 \\
    \midrule
    \bottomrule
  \end{tabular}}
  \vspace{-0.5em}
  \caption{\label{tab:error-results} \textbf{Grouped performance on \textsc{VDiff-Bench} (\%).} ``Pair macro'' indicates accuracy averaged over unique image pairs. Group columns report question-level accuracy for semantic, low-level, and OCR changes. Error columns give rates across all questions of selecting no difference, either hard negative, or an invalid response; lower is better.}
  \vspace{-1em}
\end{table}
\vspace{-0.5em}

\subsection{Group-Level Performance and Error Modes}
\label{sec:errors}
Table~\ref{tab:error-results} and Figure~\ref{fig:group-performance} aggregate model performance into higher-level semantic, low-level, and OCR change groups.
\vspace{-0.5em}
\paragraph{Semantic and low-level VDI can dissociate sharply.}
We observe that Grok and the three 7--8B models exhibit remarkable semantic--low-level gaps of 37.2--53.8 percentage points. 
Thus, success on semantic edits does not reliably predict sensitivity to changes in visual appearance, and overall accuracy can conceal qualitatively different capabilities.

\vspace{-0.5em}
\paragraph{Low-level failures primarily reflect missed changes.}
For the 4 models with the largest semantic--low-level gaps, 48.1--80.9\% of low-level questions are answered with the no-difference option, even though every image pair contains a real change. These rates substantially exceed their selection of alternative descriptions, indicating that the dominant failure is often detecting that a subtle change occurred, rather than distinguishing among competing descriptions of it.


\section{Conclusion}
We introduced \textbf{\textsc{VDiff-Bench}}, a diagnostic benchmark for fine-grained visual difference identification in MLLMs. 
The benchmark contains 1,756 four-choice questions derived from 1,543 image pairs and spans ten semantic, textual, and low-level change categories. 
By pairing each ground-truth difference with two plausible hard negatives and a no-difference distractor, \textsc{VDiff-Bench} enables deterministic scoring and interpretable error analysis.
Across 11 contemporary MLLMs, accuracy ranges from 35.8\% to 89.6\%, with no model dominating every change category. Most notably, strong semantic comparison does not guarantee low-level sensitivity: Grok and the three 7--8B models exhibit semantic--low-level gaps of 37.2--53.8 percentage points and frequently fail to register that any change occurred. Conversely, the strong performance of the Kimi models argues against a simple open- versus closed-source explanation. Together, these findings suggest that fine-grained comparative perception remains a challenging task for MLLMs, yet model capacity alone is not the only reason behind this bottleneck. 

\clearpage
\bibliography{iclr2026_conference}
\bibliographystyle{iclr2026_conference}

\clearpage
\appendix

\section{Additional Details on Dataset Construction}
\label{app:reproducibility}

\subsection{Source Composition}
\label{app:sources}
\vspace{-0.5em}
Table~\ref{tab:source-composition} reports both questions and distinct image pairs.
The distinction matters for sources whose pairs contain multiple annotated changes: Spot-the-Diff contributes 108 questions from 55 pairs, MotionEdit contributes 137 from 67, and the kids-games source contributes 102 from 12.

\begin{table}[h]
  \centering
  \small
  \vspace{-0.5em}
  \begin{tabular}{lrrl}
    \toprule
    \textbf{Source subset} & \textbf{Questions} & \textbf{Pairs} & \textbf{Primary category/categories} \\
    \midrule
    Spot-the-Diff & 108 & 55 & appearance, position \\
    MotionEdit & 137 & 67 & motion \\
    OmniEdit color & 137 & 137 & regional color \\
    OmniEdit position & 120 & 120 & position \\
    Custom illumination & 44 & 44 & illumination \\
    Kids games & 102 & 12 & mixed puzzle differences \\
    JustFamilyFun puzzles & 56 & 56 & mixed puzzle differences \\
    Kids puzzles & 34 & 34 & mixed puzzle differences \\
    Gaussian perturbation & 150 & 150 & noise/resolution \\
    Smoothing transformation & 150 & 150 & texture \\
    RGB shift & 147 & 147 & whole-image color \\
    OCR/text & 221 & 221 & OCR/text \\
    OmniDiff illumination & 151 & 151 & illumination \\
    OmniDiff substitution/size & 199 & 199 & substitution/size \\
    \midrule
    \textbf{Total} & \textbf{1,756} & \textbf{1,543} & \\
    \bottomrule
  \end{tabular}
  \caption{Source composition of the current benchmark inventory. Source names are reported for provenance; main-paper coverage is summarized by change category.}
  \label{tab:source-composition}
  \vspace{-1em}
\end{table}

\subsection{Augmentation with Programmatic Visual Changes}
\label{app:programmatic-data}

\paragraph{Adding Gaussian Noise}
We perform the first type of low-level visual transformation on 150 FFHQ images~\citep{karras2019style} by adding Gaussian noise.
For each image, we independently sample additive noise for every pixel and RGB channel from a zero-mean Gaussian distribution with standard deviation 15 in 8-bit pixel space, and clip the result to $[0,255]$. This yields global low-level corruption while preserving the image content and layout.

\paragraph{Changing RGB}
The second type of RGB-shift transformation starts from the same 150-image FFHQ image pool. 
As a first augmentation step, we apply a stronger global RGB perturbation to each image. 
As a second step, we compute the mean per-channel difference between this perturbed image and the original, remove its average channel offset to isolate chromatic rather than brightness change, scale the resulting color direction by 0.55, and cap the maximum absolute channel offset at 12 pixel values before applying it to the final image. 
This yields subtle global tone shifts while keeping the transformed image close to the original.
The final benchmark retains 147 examples after excluding three construction-time flagged images.

\paragraph{Changing Texture / Smoothing}
The texture transformation / smoothing transformation uses the same 150-image FFHQ image pool. 
We apply a skin- and edge-aware smoothing transform: the image is blurred with a Gaussian kernel whose standard deviation is sampled from $[2.4,3.6]$, while a soft skin mask, estimated from YCbCr color thresholds, increases smoothing on skin-like regions. 
Edges are protected using a luminance high-pass signal, and residual detail is mixed back with a random scale in $[0.18,0.36]$.
After smoothing, we add a small luminance-dependent highlight term to bright skin-like regions, reduce saturation by mixing the image toward grayscale, lower contrast around the mid-gray point, and apply a gamma correction sampled close to one. 
Together, these operations suppress fine texture and create a smoother, plastic-like appearance while preserving the image layout and object identities.

\paragraph{Changing Illumination}
We augment the OmniDiff~\citep{liu2025omnidiff} illumination subset by constructing images with more subtle illumination changes.
We use the first 50 sorted images from the Country211 test split~\citep{radford2021clip} as the seed split.
Each image is adjusted by a gamma-plus-linear transform,
$\mathrm{clip}(255 \cdot x^\gamma + b)$, where $x$ is the normalized RGB image. 
For brighter examples, $\gamma \sim U(0.70,0.85)$ and $b \sim \{5,\ldots,15\}$; for darker examples, $\gamma \sim U(1.15,1.30)$ and $b \sim \{-15,\ldots,-5\}$.

\subsection{Augmentation with Image Editing Differences}
\label{app:construction}
We additionally augment the OCR change and object position change categories in our evaluation data.
Inspired by the recent advancement in high-performance image editing models, we utilize frontier commercial image editing model to help with this data augmentation process.

\paragraph{Changing OCR Text}
The OCR subset contains 221 examples: 50 existing OCR/text-related pairs imported from OmniDiff, and 171 custom text edits initialized from MLT19 and TextOCR images.
For the custom OCR edits, we select readable, approximately horizontal text regions with moderate size, avoiding boxes smaller than $50 \times 18$ pixels or covering more than 40\% of the image. 
We propose one of four localized textual edits: replacing one same-type character, replacing one digit, swapping adjacent characters, or replacing a word with another same-script word visible in the image. 
The selected crop is padded by 24 pixels, and then edited with \textit{gemini-3-pro-image} which is instructed to also preserve font, color, perspective, background, lighting, and texture, and pasted back with an 8-pixel feathered boundary. 
We then validate that the target text appears and that non-text scene content remains unchanged.
Below, we provide the prompt for the image editing model on the OCR edit task.

\begin{tcolorbox}[
    enhanced,
    colback=white,
    colframe=gray!70!black,
    coltitle=white,
    colbacktitle=gray!70!black,
    title={\textbf{OCR Edit Prompt}},
    fonttitle=\bfseries\normalsize,
    boxrule=1.2pt,
    arc=3mm,
    outer arc=3mm,
    left=8mm,
    right=8mm,
    top=3mm,
    bottom=3mm,
    titlerule=0pt,
    toptitle=1.5mm,
    bottomtitle=1.5mm
]
\normalsize
You are looking at a crop from a scene image that contains text in \texttt{\{script\}} script.

Change the text ``\texttt{\{orig\_\_text\}}'' to ``\texttt{\{new\_\_text\}}''.

Match the original font, color, size, alignment, perspective, and texture exactly.

Keep the background, lighting, and every other element pixel-identical to the input.

Only the text changes. Output the edited image.
\end{tcolorbox}

\paragraph{Changing Object Position}
We further augment the ``position change'' category in the evaluation data.
We take 150 seed images from the ``swap'' task of the OmniDiff dataset to begin with; the reason for selecting the ``swap'' images is because the nature of this task guarantees at least 2 clearly visible objects in images, whose positions can be cleanly changed.
For each seed image, we first utilize \textit{gpt-5.4-mini} to proposes 5 candidate edits, each specifying a visible, countable object, its original location, a plausible new location, an editing instruction, and a reference difference statement beginning with ``In the second image,''. 
The prompt explicitly disallows adding, removing, recoloring, resizing, rotating, duplicating, replacing, or deforming objects, so the intended change is restricted to spatial relocation. 
We use the first candidate by default and append preservation constraints requiring the same scene, camera angle, lighting, style, object identities, colors, and object counts. 
The selected instruction is then passed to \textit{gpt-image-2}~\citep{openai2026gptimage2} to produce an edited image.
Below, we provide the prompts for the MLLM to propose potential positional edits, as well as the final image editing prompt template.

\begin{tcolorbox}[
    enhanced,
    colback=white,
    colframe=gray!70!black,
    coltitle=white,
    colbacktitle=gray!70!black,
    title={\textbf{Position Edit Proposal Prompt}},
    fonttitle=\bfseries\normalsize,
    boxrule=1.2pt,
    arc=3mm,
    outer arc=3mm,
    left=8mm,
    right=8mm,
    top=3mm,
    bottom=3mm,
    titlerule=0pt,
    toptitle=1.5mm,
    bottomtitle=1.5mm
]
\normalsize
You are designing image-edit prompts for a rigorous visual difference benchmark.\\
Look at the image and propose exactly 5 candidate edits where ONE visible object changes position.\\
Each candidate must move an existing visible object to a different plausible location in the same scene.\\
Do not add, remove, recolor, resize, restyle, rotate, deform, duplicate, or replace objects.\\
Prefer clearly localized, countable objects whose position change would be easy to verify.\\
Avoid vague background regions.\\
The \texttt{edit\_prompt} must be clear and effective to instruct an image editing model to preserve everything except that single object's position.\\
The \texttt{difference\_description} must be a short dataset annotation beginning with 'In the second image,'.\\

Original OmniEdit swap instruction for context only: \\
- \texttt{\{original\_instruction\_1\}}\\
- \texttt{\{original\_instruction\_2\}}\\

\end{tcolorbox}

\begin{tcolorbox}[
    enhanced,
    colback=white,
    colframe=gray!70!black,
    coltitle=white,
    colbacktitle=gray!70!black,
    title={\textbf{Position Edit Prompt}},
    fonttitle=\bfseries\normalsize,
    boxrule=1.2pt,
    arc=3mm,
    outer arc=3mm,
    left=8mm,
    right=8mm,
    top=3mm,
    bottom=3mm,
    titlerule=0pt,
    toptitle=1.5mm,
    bottomtitle=1.5mm
]
\normalsize
\texttt{\{selected\_\_candidate\_\_edit\_\_prompt\}}\\

Keep the same scene, camera angle, lighting, style, object identities, colors, and object counts. Only change the selected object's position.
\end{tcolorbox}


\subsection{Additional Details on False Differences Generation}
We prompt the \textit{Gemini-2.5-pro} model to inspect each image pair data and construct false differences that are semantically plausible but factually non-existent.
This false difference candidates are then used as distractor options in the final multiple choice question construction, to challenge the evaluated models.

\paragraph{Generation Prompt.}
For each image pair, the model receives a list of ground truth differences in them and are prompted to use these as references to create 2 to 6 statements that are factually incorrect but visually plausible, formatted similarly as the references.
The prompt instructs the model to consider 4 strategies: swapping the target object, swapping the source object, changing an attribute of the target, and introducing a plausible alternative change to an entity visible in at least one image.
The prompt also explicitly prohibits invented entities, indirect references to the true change, and complex relational wording.
The prompt additionally requests up to five potentially missing true differences for audit; these suggestions are not treated as annotations, but any overlap with a displayed candidate alternative is conservatively flagged for visual adjudication.
Fig. \ref{fig:negative-generation-prompt} below presents the prompt for the model to generate the false differences.

\begin{tcolorbox}[
    enhanced,
    breakable,
    colback=white,
    colframe=gray!70!black,
    coltitle=white,
    colbacktitle=gray!70!black,
    title={\textbf{False Differences Generation Prompt}},
    fonttitle=\bfseries\normalsize,
    boxrule=1.2pt,
    arc=3mm,
    outer arc=3mm,
    left=8mm,
    right=8mm,
    top=3mm,
    bottom=3mm,
    titlerule=0pt,
    toptitle=1.5mm,
    bottomtitle=1.5mm
]
\normalsize
\small

\noindent\textbf{\#\#\# Role}

You are a precision image-analysis engine specialized in generating hard-negative distractors for change detection datasets.

\medskip
\noindent\textbf{\#\#\# Context}

Dataset Category: \texttt{\{category\_type\}}

Expected Change Logic: \texttt{\{allowed\_family\_text\}}

\medskip
\noindent\textbf{\#\#\# Input Data}

1. Images: [Provided]

2. Ground Truth (GT) Differences:

\texttt{\{truth\_bullets\}}

\medskip
\noindent\textbf{\#\#\# Task 1: Generate False Differences (Hard Negatives)}

Create \texttt{\{distractor\_count\}} to \texttt{\{max\_false\_differences\}} statements that are FACTUALLY INCORRECT but visually plausible based on the scene.

Each false difference should be phrased and formatted just like the GT.

Use these strategies:

\begin{itemize}
    \item \textbf{GT Swap:} Take a GT change and swap the ending object, e.g., B in ``A is changed to B'', to another object that is also present in the image, preferably located close to A / B.

    \item \textbf{Input Swap:} Take a GT change and swap the beginning object, e.g., A in ``A is changed to B'', to another object that is visible in the image, preferably located close to A / B.

    \item \textbf{GT change:} Take a GT change and change characteristics of the ending object, e.g., \texttt{adj\_Y B} in ``\texttt{adj\_X A} is changed to a \texttt{adj\_Y B}'', to another characteristic, e.g., ``\texttt{adj\_X A} is changed to a \texttt{adj\_P B}'', such as color, material, shape, etc.

    \item \textbf{Non-Substitution changes:} Randomly add false changes from non-substitution categories, such as color changes, left/right/closer/further movement, character or shape changes, illumination changes, noise changes, or RGB tone changes. Make sure the object you mention exists in at least one of the images.
\end{itemize}

\noindent\textbf{Constraints:}

\begin{itemize}
    \item Your false differences should be phrased and formatted just like the GT differences.

    \item The subject of your false statement MUST exist and be visible in at least one of the images. Do not invent new objects.

    \item Preserve the main changed object from the GT in at least one but at most two generated false differences.

    \item Do NOT mention the ground truth. For example, if the ground truth is ``the white SUV has disappeared'', do NOT describe something else as moved relative to ``where the SUV was''.

    \item Do NOT use complex relational phrasing such as ``behind where X was'', ``next to where Y had been'', or ``in the spot formerly occupied by''.

    \item Do NOT generate more than two GT Swap differences that look similar to each other.
\end{itemize}

\medskip
\noindent\textbf{\#\#\# Task 2: Identify Potential Missing True Differences}

List up to 5 real visual changes visible in the images that were NOT captured in the provided Ground Truth list.

If the GT is exhaustive, return an empty list.

\medskip
\noindent\textbf{\#\#\# Output Format (JSON)}

Return the response in this exact JSON structure:

\begin{verbatim}
{
  "false_differences": [
    {
      "statement": "The false sentence here.",
      "based_on_true_difference": "The exact GT sentence this is mimicking.",
      "strategy_used": "Object Swap / Spatial Error / State Error"
    }
  ],
  "potential_missing_true_differences": ["Statement 1", "Statement 2"]
}
\end{verbatim}
\end{tcolorbox}
\captionof{figure}{Prompt used to generate hard-negative distractors.}
\label{fig:negative-generation-prompt}

\section{Evaluation Prompt}
\label{app:prompt}
All main evaluations use the following textual template, with the four options substituted from the corresponding per-example inference artifact:

\begin{tcolorbox}[
    enhanced,
    colback=white,
    colframe=gray!70!black,
    coltitle=white,
    colbacktitle=gray!70!black,
    title={\textbf{Evaluation Prompt}},
    fonttitle=\bfseries\normalsize,
    boxrule=1.2pt,
    arc=3mm,
    outer arc=3mm,
    left=8mm,
    right=8mm,
    top=3mm,
    bottom=3mm,
    titlerule=0pt,
    toptitle=1.5mm,
    bottomtitle=1.5mm
]
\normalsize
You are given image A and image B.\\[3pt]
Choose the option that correctly describes a real difference between image A and image B.\\
Options:\\
A. [option A]\\
B. [option B]\\
C. [option C]\\
D. [option D]\\[3pt]
Answer ONLY one of the choice labels (e.g. `A.') without other text:
\end{tcolorbox}


Images are supplied directly to the model in A--B order; no difference mask is used.
All runs use temperature zero; Qwen3-VL-8B Thinking is run with reasoning disabled, and no separate reasoning trace is requested or recorded for the remaining models.


\section{Additional Results}

\subsection{No-Difference Selection on Low-Level Changes}
As shown in Table~\ref{tab:no-diff-rates}, models differ substantially in how often they incorrectly judge an image pair as unchanged. 
Gemini 3.5 Flash, Gemini 3.1 Pro, GPT-5.4, and Gemini 2.5 Flash select the no-difference option on only 0.5\%, 1.6\%, 5.2\%, and 12.1\% of all questions, respectively, whereas this rate rises to 24.8--43.1\% for the smaller models. 
Missed-change selections are particularly frequent for the low-level visual changes---including whole-image RGB, texture, and noise differences---exceeding 70\% in several cases.
This shows that models still fail at correctly capturing visual differences that are not straightforwardly observable.

\begin{table}[h]
  \centering
  \small
  \begin{tabular}{lrrrrr}
    \toprule
    \textbf{Model} & \textbf{Whole color} & \textbf{Noise} & \textbf{Texture} & \textbf{Illum.} & \textbf{Overall} \\
    \midrule
    GPT-5.4 & 9.5 & 0.0 & 9.6 & 14.4 & 5.2 \\
    Gemini 2.5 Flash & 24.5 & 0.0 & 39.0 & 21.0 & 12.1 \\
    Gemini 3.1 Pro & 1.4 & 0.0 & 2.3 & 4.6 & 1.6 \\
    Gemini 3.5 Flash & 2.0 & 0.0 & 0.0 & 1.0 & 0.5 \\
    Grok 4.3 & 4.1 & 79.3 & 72.3 & 35.4 & 21.8 \\
    Doubao Seed 1.6 Vision & 0.0 & 0.0 & 4.0 & 6.2 & 1.3 \\
    Qwen3-VL-8B Thinking & 74.8 & 22.0 & 74.0 & 35.4 & 24.8 \\
    InternVL3.5-8B & 93.2 & 87.3 & 90.4 & 42.1 & 36.8 \\
    LLaVA-OneVision-Qwen2-7B & 93.9 & 88.7 & 72.3 & 72.8 & 43.1 \\
    Kimi K2.5 & 0.0 & 0.0 & 9.0 & 9.7 & 3.4 \\
    Kimi K3 & 15.0 & 0.0 & 10.7 & 15.9 & 5.9 \\
    \bottomrule
  \end{tabular}
  \caption{Rate (\%) at which each model selects the no-difference distractor. The first four columns are category-specific; ``Overall'' covers all 1,756 questions. High rates indicate a missed-change selection rather than confusion with another detailed description.}
  \label{tab:no-diff-rates}
  \vspace{-0.8em}
\end{table}

\subsection{Representative Failure Cases}
\label{app:failures}

\begin{figure}[htbp]
  \centering
  \begin{minipage}{0.31\textwidth}
    \centering
    \includegraphics[width=0.48\linewidth]{figs/failure_texture_a.png}\hfill
    \includegraphics[width=0.48\linewidth]{figs/failure_texture_b.png}\\[-2pt]
    \small (a) Subtle smoothing
  \end{minipage}\hfill
  \begin{minipage}{0.31\textwidth}
    \centering
    \includegraphics[width=0.48\linewidth]{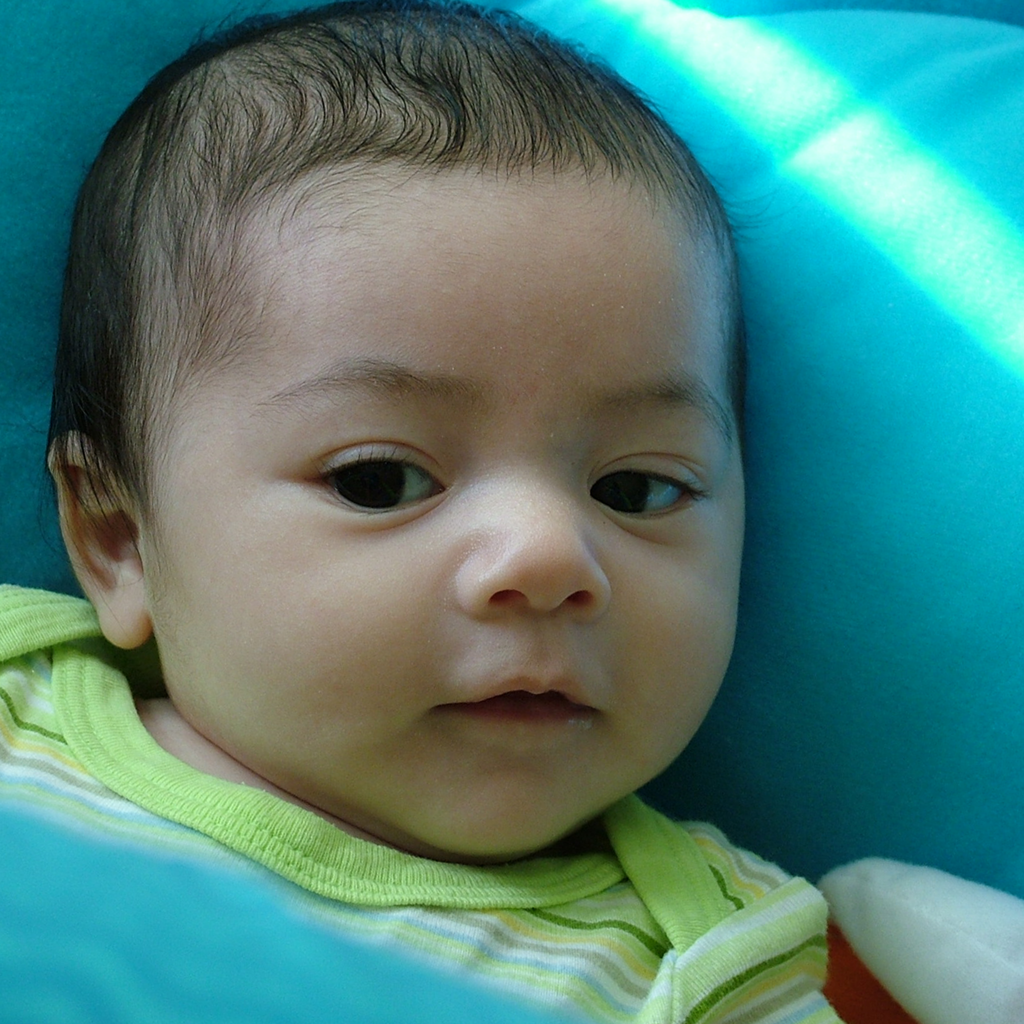}\hfill
    \includegraphics[width=0.48\linewidth]{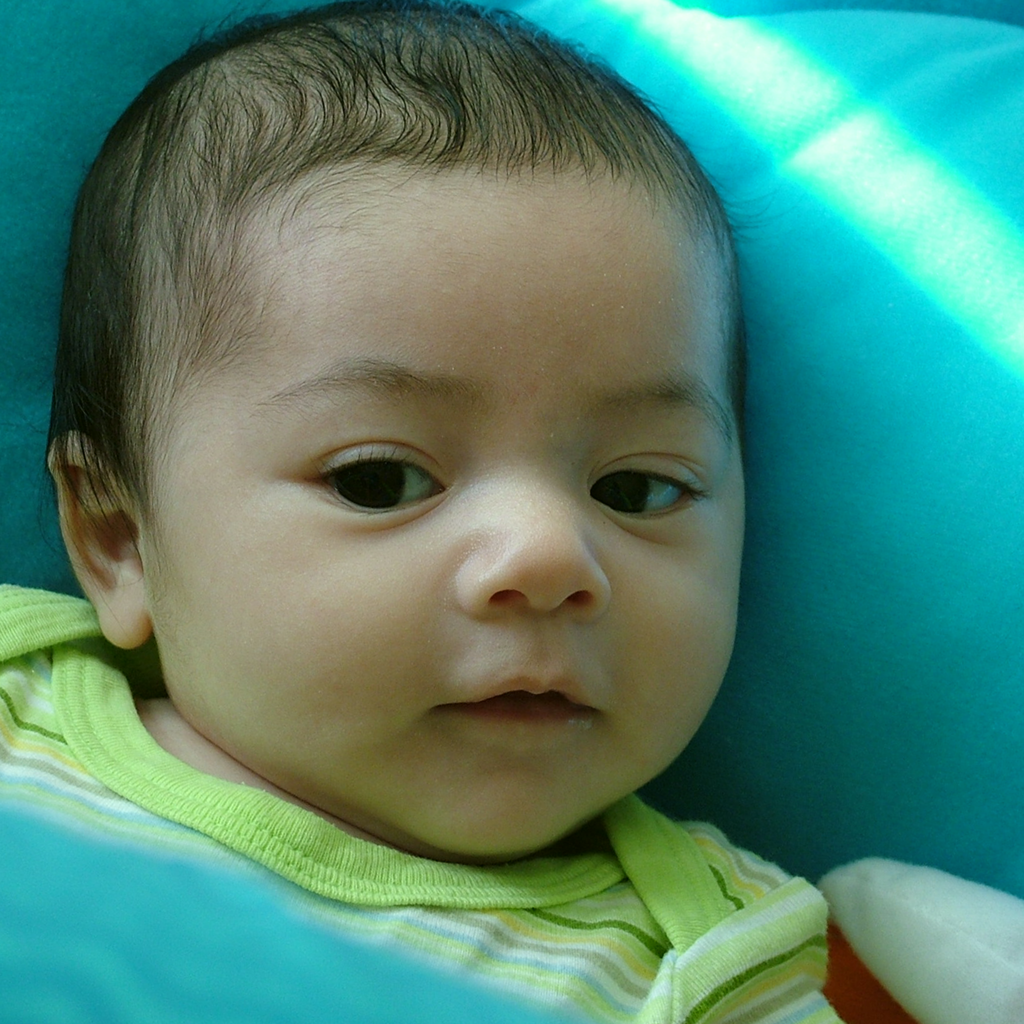}\\[-2pt]
    \small (b) Global RGB shift
  \end{minipage}\hfill
  \begin{minipage}{0.31\textwidth}
    \centering
    \includegraphics[width=0.48\linewidth]{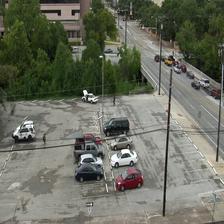}\hfill
    \includegraphics[width=0.48\linewidth]{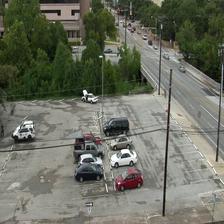}\\[-2pt]
    \small (c) Two people disappear
  \end{minipage}
  \caption{Representative no-difference selections. (a) Smoothing: Qwen, Gemini, LLaVA, and InternVL select no difference; GPT-5.4 selects the keyed answer. (b) Global RGB shift: Qwen, LLaVA, and InternVL select no difference. (c) Two people disappear: LLaVA and InternVL select no difference.}
  \label{fig:failure-cases}
\end{figure}

Figure~\ref{fig:failure-cases} illustrates three distinct sources of difficulty.
The smoothing operation changes local texture without altering scene semantics; the RGB shift distributes evidence across the entire image; and the surveillance example requires locating small missing entities in a low-resolution frame.
The repeated no-difference choice is consistent with the aggregate behavior in Table~\ref{tab:no-diff-rates}.

\FloatBarrier

\subsection{Additional Qualitative Results}
\label{appendix:additional-qualitative-results}
Below, we provide more qualitative results on how different models perform on data from each category of our benchmark.
\begingroup
\definecolor{VDPosition}{HTML}{E97943}
\definecolor{VDMotion}{HTML}{E95D63}
\definecolor{VDRegional}{HTML}{DC4F8C}
\definecolor{VDWhole}{HTML}{BE55B5}
\definecolor{VDAppear}{HTML}{965BC8}
\definecolor{VDNoise}{HTML}{7468D0}
\definecolor{VDTexture}{HTML}{5676CC}
\definecolor{VDSubstitution}{HTML}{3D8BCB}
\definecolor{VDOCR}{HTML}{229FC4}
\definecolor{VDIllumination}{HTML}{149BAD}
\definecolor{VDGreen}{HTML}{16865C}
\definecolor{VDRed}{HTML}{CC4054}

\newcommand{\VDImageHeight}{1.35in}
\newcommand{\VDPair}[2]{%
  \makebox[\linewidth][c]{%
    \begin{tikzpicture}[baseline=(current bounding box.center)]
      \node[inner sep=0,anchor=west] (A) at (0,0)
        {\includegraphics[width=0.49\linewidth,height=\VDImageHeight,keepaspectratio]{#1}};
      \node[inner sep=0,anchor=west] (B) at (A.east)
        {\includegraphics[width=0.49\linewidth,height=\VDImageHeight,keepaspectratio]{#2}};
      \node[anchor=north west,inner sep=1.2pt,fill=black!78,text=white,font=\bfseries\tiny]
        at (A.north west) {A};
      \node[anchor=north west,inner sep=1.2pt,fill=black!78,text=white,font=\bfseries\tiny]
        at (B.north west) {B};
    \end{tikzpicture}%
  }%
}

\newcommand{\VDChoice}[2]{%
  \par\noindent\hangindent=1.25em\hangafter=1\textbf{#1:} #2%
}
\newcommand{\VDTrue}[3]{%
  \par\noindent\hangindent=1.25em\hangafter=1%
  \textcolor{#1}{\textbf{#2: #3}}%
}
\newcommand{\VDYes}{\textcolor{VDGreen}{\ding{51}}}
\newcommand{\VDNo}{\textcolor{VDRed}{\ding{55}}}
\newcommand{\VDModel}[3]{%
  \raisebox{-0.25ex}{\includegraphics[height=0.10in]{figs/#1}}%
  {\fontsize{4.2}{4.2}\selectfont\bfseries #2}\,#3%
}
\newcommand{\VDModels}[6]{%
  \begin{tabular*}{\linewidth}{@{}>{\bfseries\color{black!55}}l@{\extracolsep{\fill}}cccccc@{}}
    \fontsize{5.0}{5.0}\selectfont MODELS & #1 & #2 & #3 & #4 & #5 & #6
  \end{tabular*}%
}
\newcommand{\VDLegend}{%
  \begin{tabular*}{\linewidth}{@{\extracolsep{\fill}}cccccc@{}}
    \VDModel{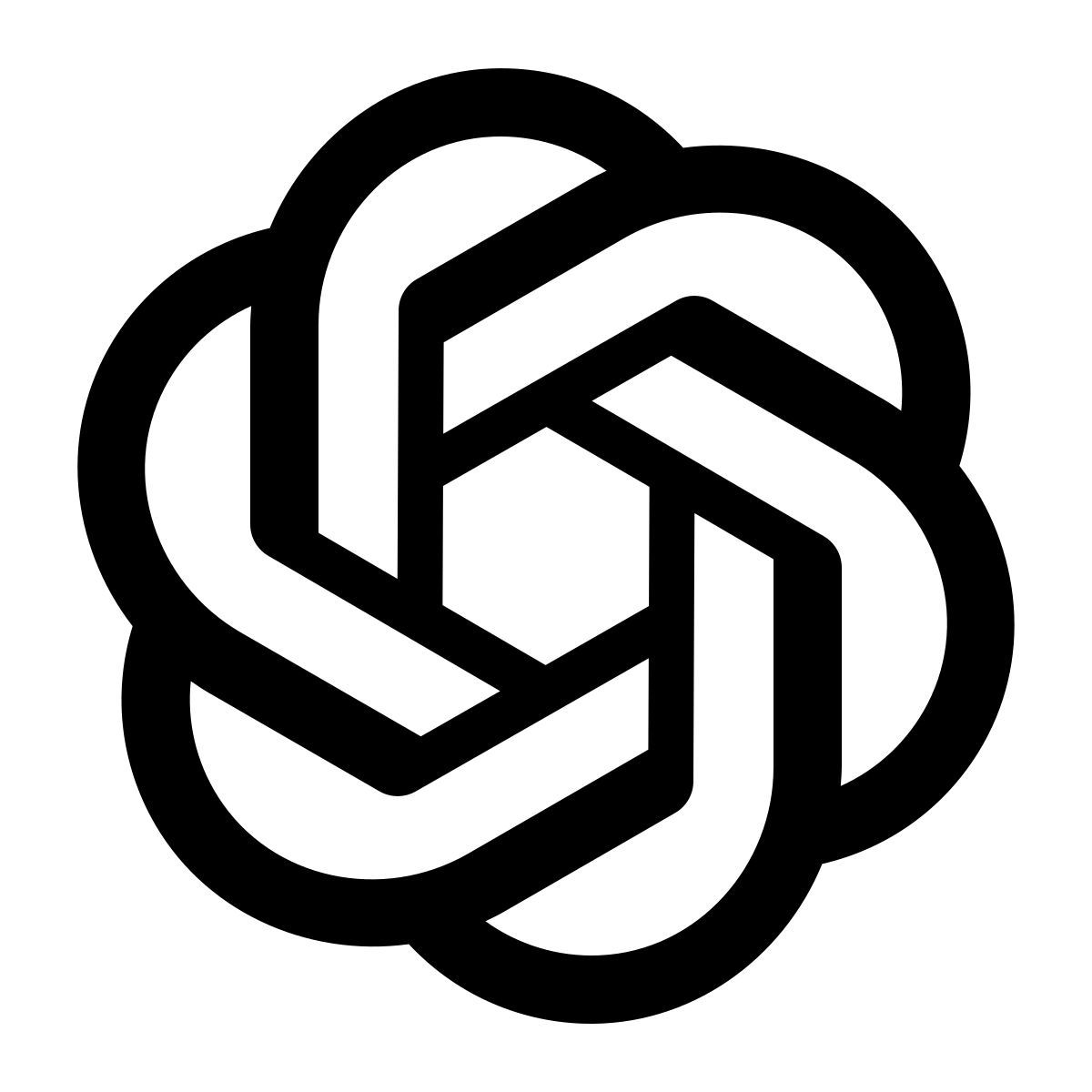}{}{GPT-5.4} &
    \VDModel{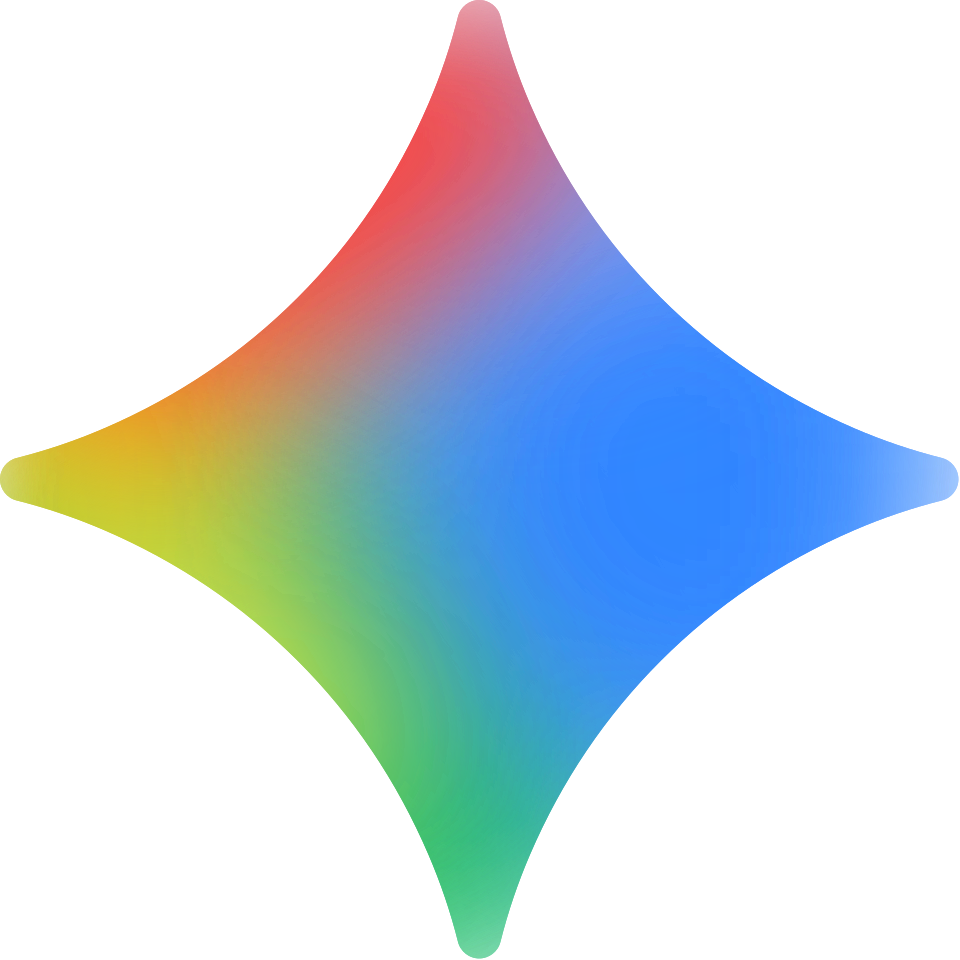}{3.1}{Gemini 3.1} &
    \VDModel{gemini.png}{3.5}{Gemini 3.5} &
    \VDModel{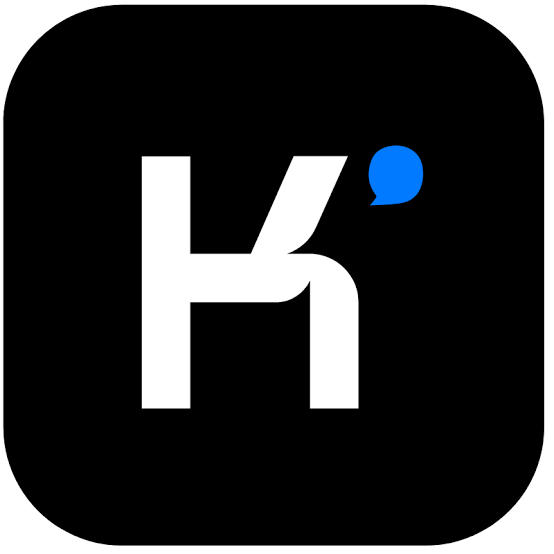}{}{Kimi K2.5} &
    \VDModel{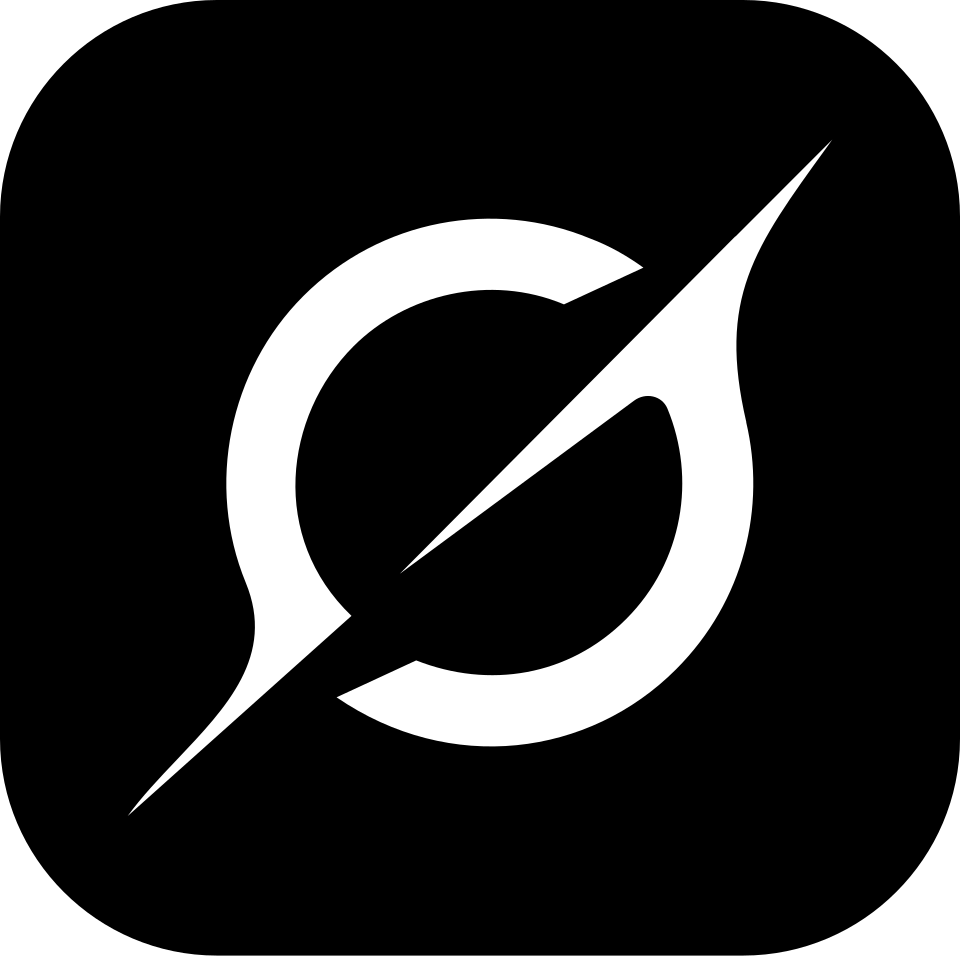}{}{Grok 4.3} &
    \VDModel{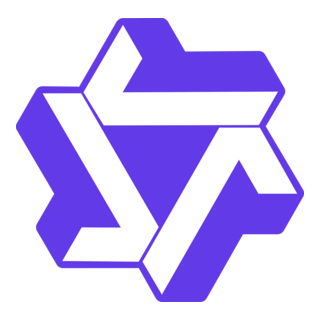}{}{Qwen3-VL-8B}
  \end{tabular*}%
}
\newcommand{\VDPanel}[6]{%
  {\color{#2}\rule{1.1mm}{1.65ex}}\hspace{0.45em}%
  {\fontsize{7.2}{7.8}\selectfont\bfseries\MakeUppercase{#1}}\par
  \vspace{0.35ex}
  \VDPair{#3}{#4}\par
  \vspace{0.25ex}
  {\fontsize{5.35}{6.15}\selectfont\raggedright #5\par}
  \vspace{0.8ex}
  {\fontsize{5.1}{5.8}\selectfont #6}\par
  \vspace{0.35ex}{\color{black!16}\hrule height 0.35pt}
  \vfill
}

\begin{figure}[!t]
  \centering
  \VDLegend
  \vspace{0.8ex}

  \begin{minipage}[t][0.245\textheight][t]{0.492\textwidth}
    \VDPanel{Position}{VDPosition}
      {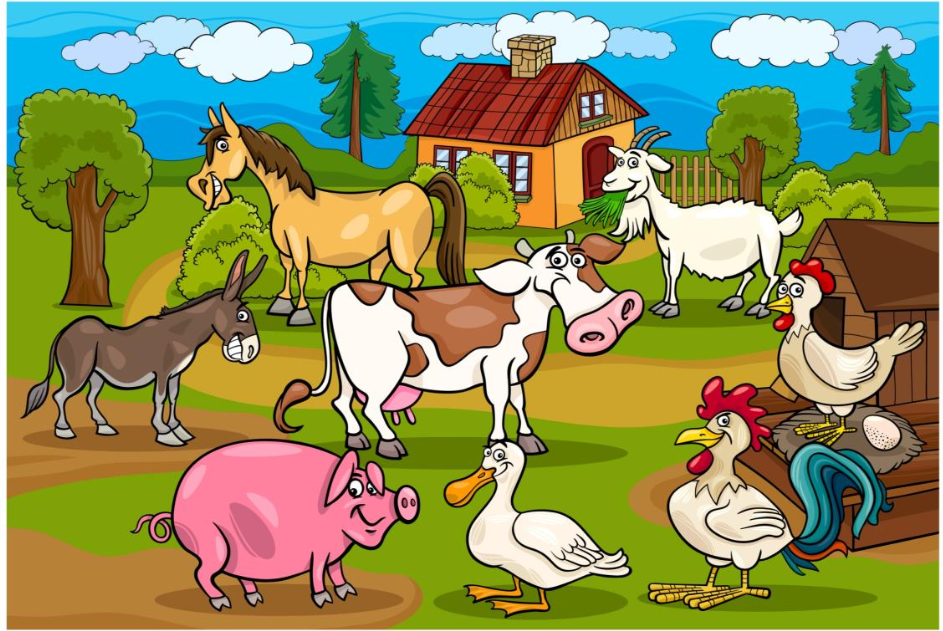}
      {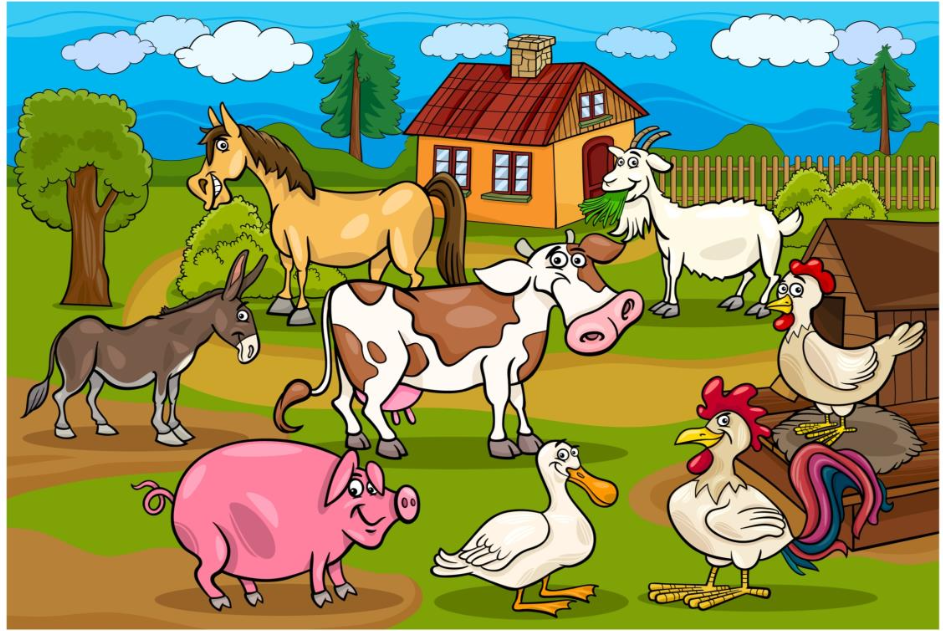}
      {\VDChoice{A}{The white duck in the center is facing left in A but facing front in B.}
       \VDTrue{VDPosition}{B}{The pig's tail points downward in A but curls upward in B.}
       \VDChoice{C}{The horse's tail changes color from brown in A to black in B.}
       \VDChoice{D}{No difference between A and B.}}
      {\VDModels
        {\VDModel{openai.png}{}{\VDYes}}
        {\VDModel{gemini.png}{3.1}{\VDNo}}
        {\VDModel{gemini.png}{3.5}{\VDNo}}
        {\VDModel{kimi.png}{}{\VDNo}}
        {\VDModel{grok.png}{}{\VDYes}}
        {\VDModel{qwen.png}{}{\VDNo}}}
  \end{minipage}\hfill
  \begin{minipage}[t][0.245\textheight][t]{0.492\textwidth}
    \VDPanel{Motion}{VDMotion}
      {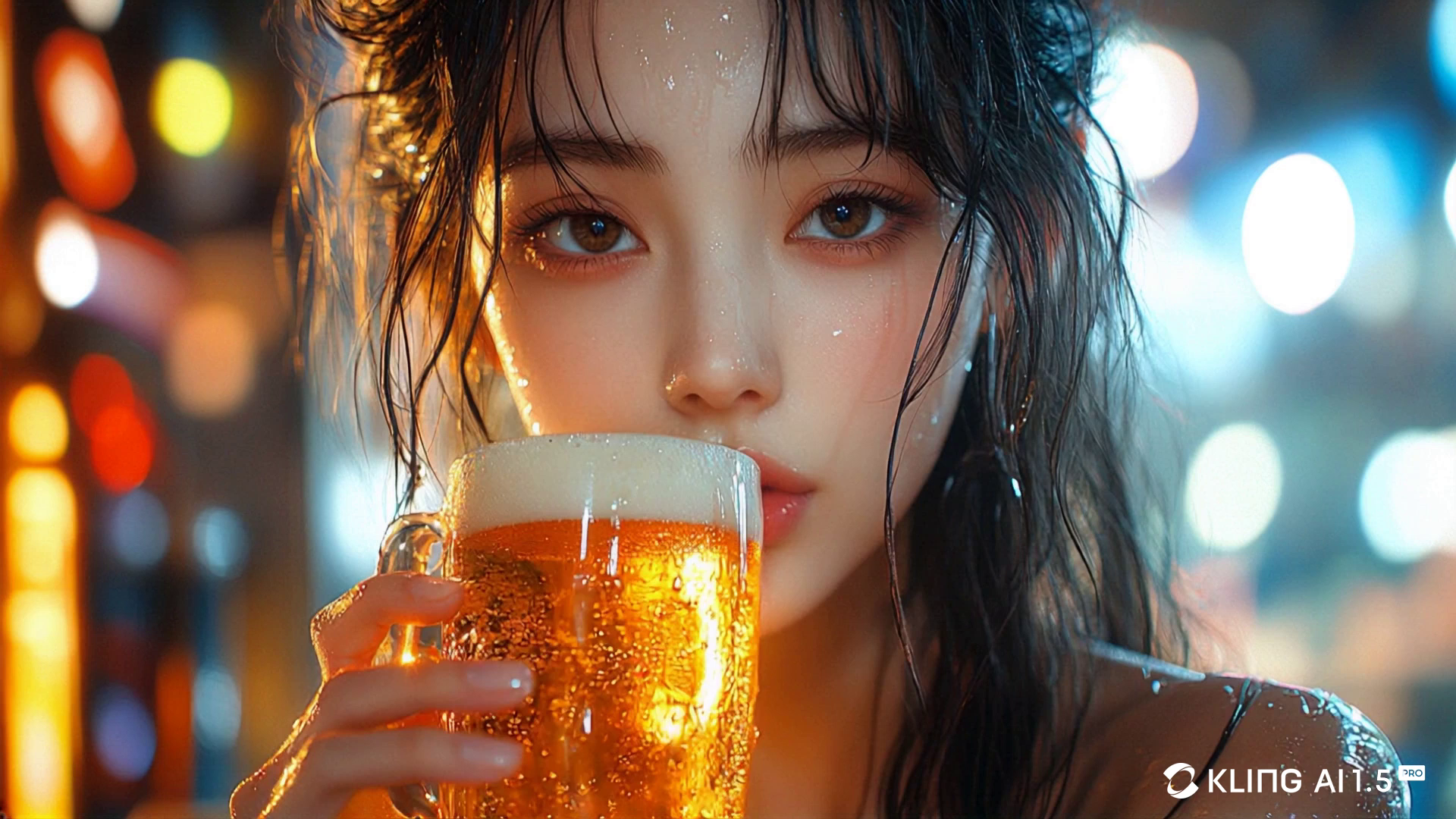}
      {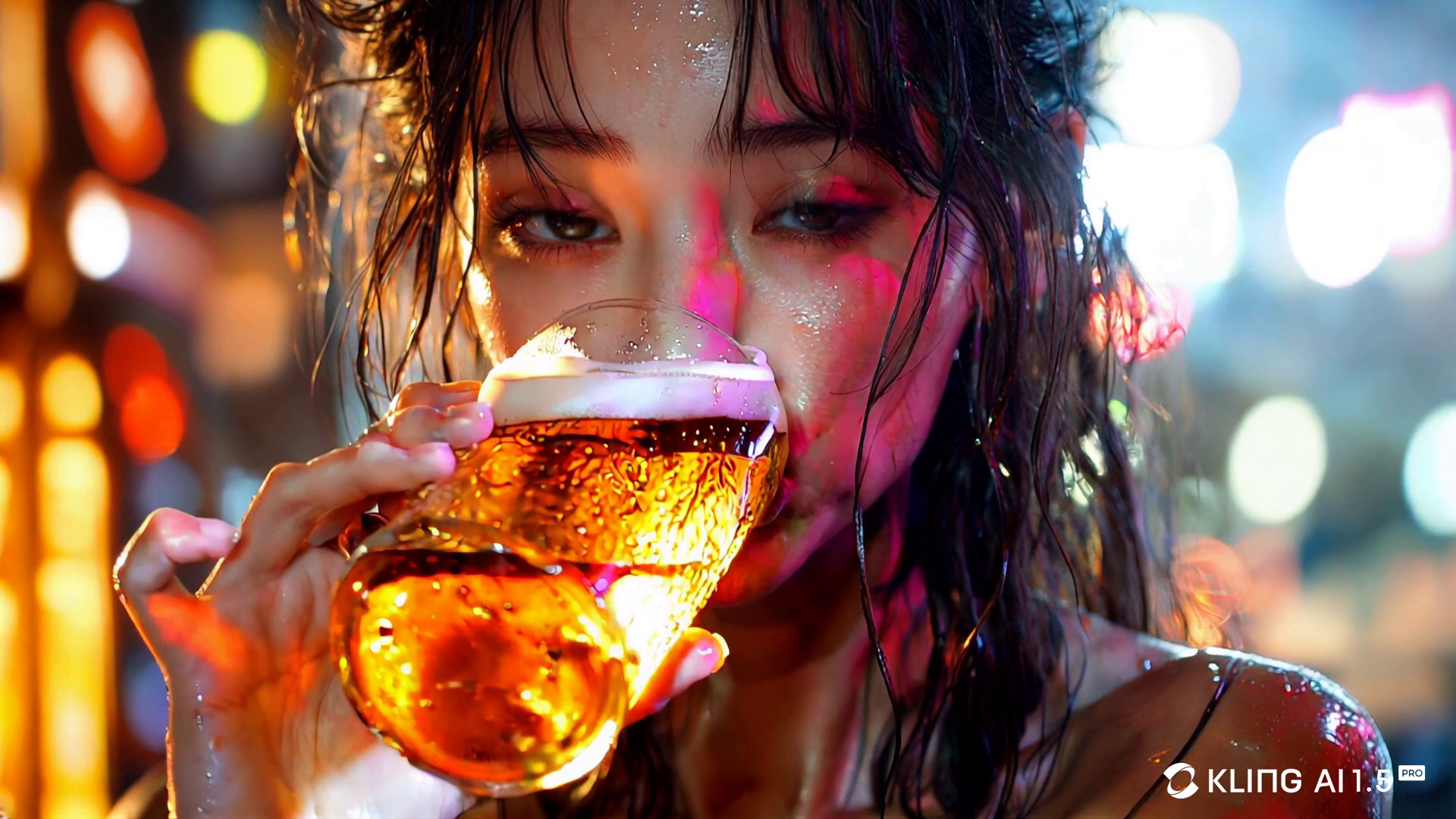}
      {\VDTrue{VDMotion}{A}{In B, the woman's eyes are looking down.}
       \VDChoice{B}{The beer glass is now tilted and has a higher liquid level.}
       \VDChoice{C}{The woman is now drinking from the glass and looking up.}
       \VDChoice{D}{No difference between A and B.}}
      {\VDModels
        {\VDModel{openai.png}{}{\VDNo}}
        {\VDModel{gemini.png}{3.1}{\VDYes}}
        {\VDModel{gemini.png}{3.5}{\VDYes}}
        {\VDModel{kimi.png}{}{\VDNo}}
        {\VDModel{grok.png}{}{\VDYes}}
        {\VDModel{qwen.png}{}{\VDYes}}}
  \end{minipage}

  \vspace{0.8ex}
  \begin{minipage}[t][0.245\textheight][t]{0.492\textwidth}
    \VDPanel{Regional color}{VDRegional}
      {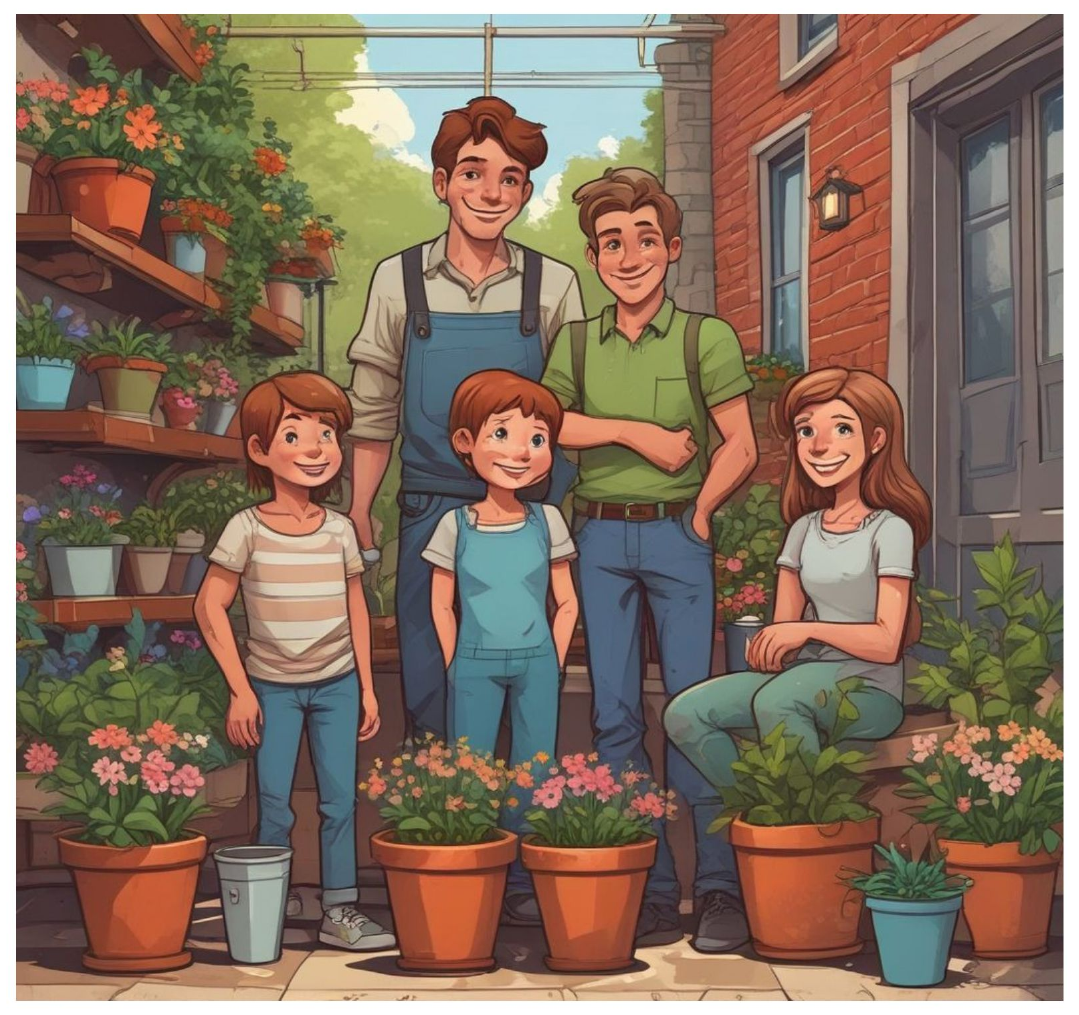}
      {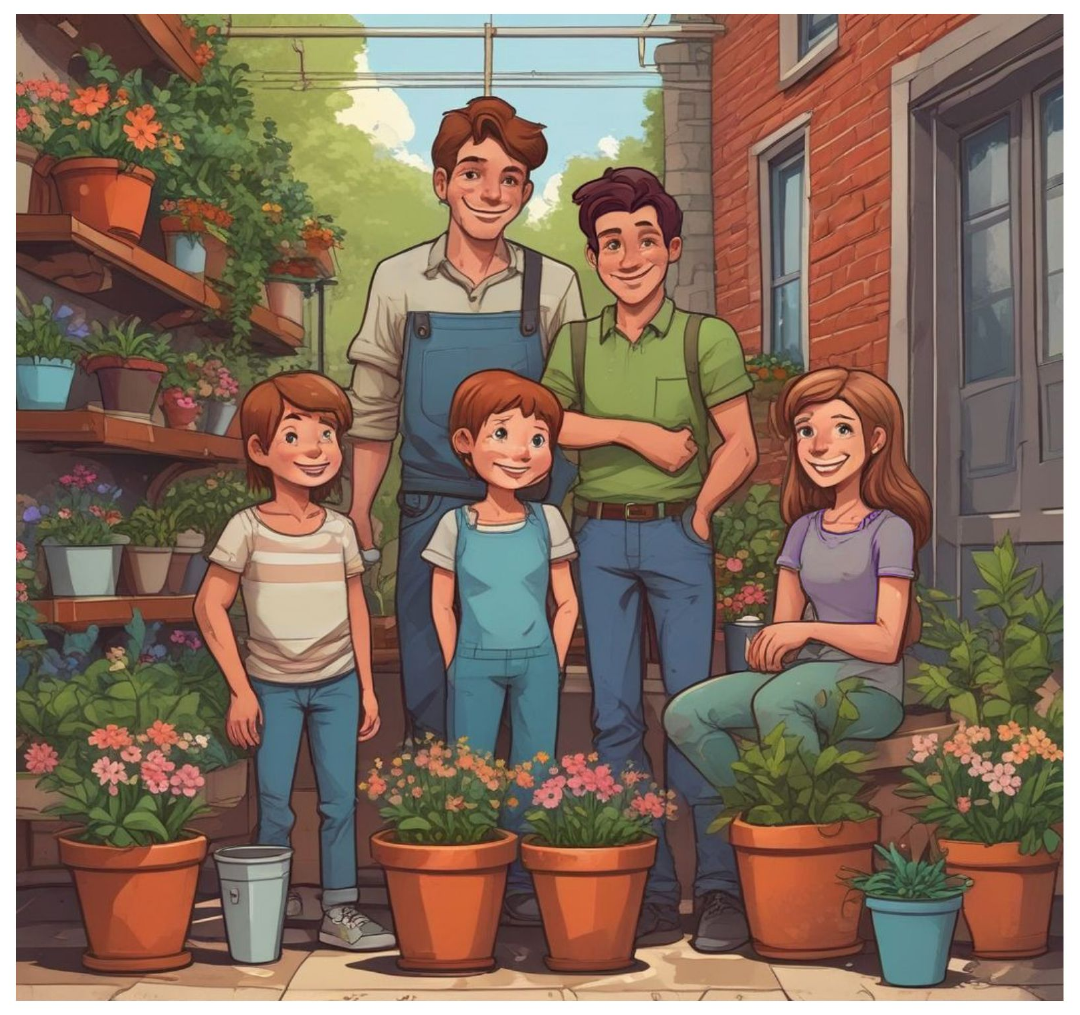}
      {\VDTrue{VDRegional}{A}{A flower pot on the middle shelf on the left side is green in A but grey in B.}
       \VDChoice{B}{No difference between A and B.}
       \VDChoice{C}{The small pot on the far right is white in A and changes to blue in B.}
       \VDChoice{D}{The overalls worn by the child in the center are light blue in A and have changed to a darker blue in B.}}
      {\VDModels
        {\VDModel{openai.png}{}{\VDNo}}
        {\VDModel{gemini.png}{3.1}{\VDYes}}
        {\VDModel{gemini.png}{3.5}{\VDNo}}
        {\VDModel{kimi.png}{}{\VDYes}}
        {\VDModel{grok.png}{}{\VDYes}}
        {\VDModel{qwen.png}{}{\VDNo}}}
  \end{minipage}\hfill
  \begin{minipage}[t][0.245\textheight][t]{0.492\textwidth}
    \VDPanel{Whole-image color}{VDWhole}
      {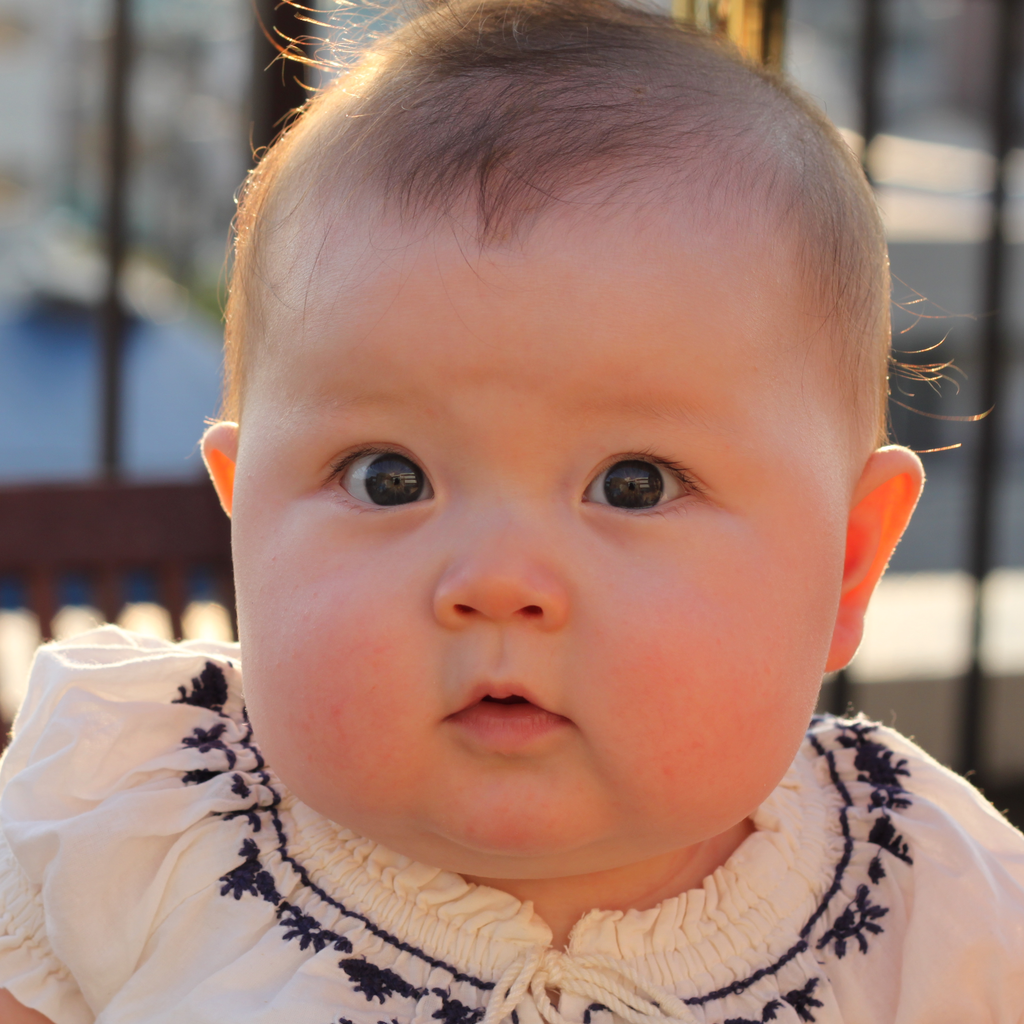}
      {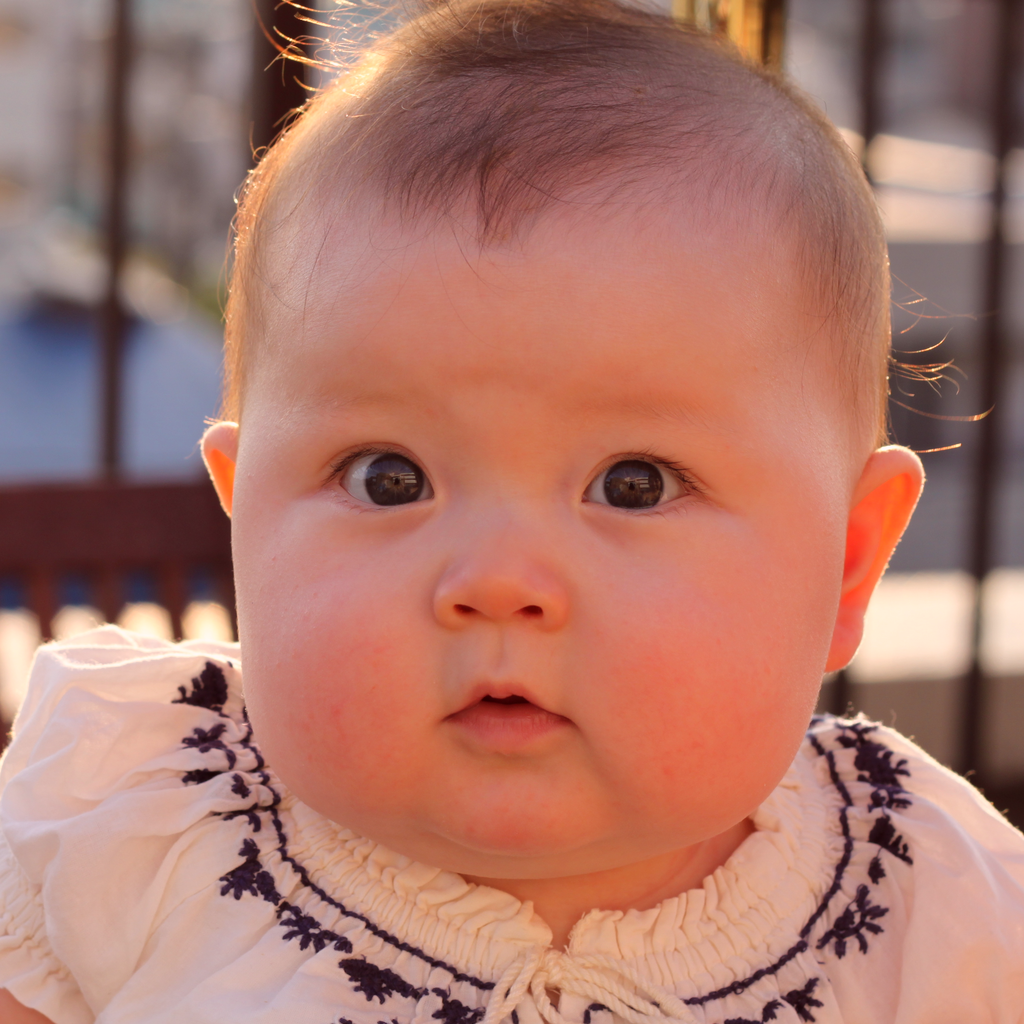}
      {\VDChoice{A}{No difference between A and B.}
       \VDChoice{B}{B has a cooler blue--green color tone than A.}
       \VDChoice{C}{The baby's facial expression is different in B.}
       \VDTrue{VDWhole}{D}{B has a warmer red--magenta color tone than A.}}
      {\VDModels
        {\VDModel{openai.png}{}{\VDNo}}
        {\VDModel{gemini.png}{3.1}{\VDYes}}
        {\VDModel{gemini.png}{3.5}{\VDYes}}
        {\VDModel{kimi.png}{}{\VDYes}}
        {\VDModel{grok.png}{}{\VDYes}}
        {\VDModel{qwen.png}{}{\VDNo}}}
  \end{minipage}

  \vspace{0.8ex}
  \begin{minipage}[t][0.245\textheight][t]{0.492\textwidth}
    \VDPanel{Appear / disappear}{VDAppear}
      {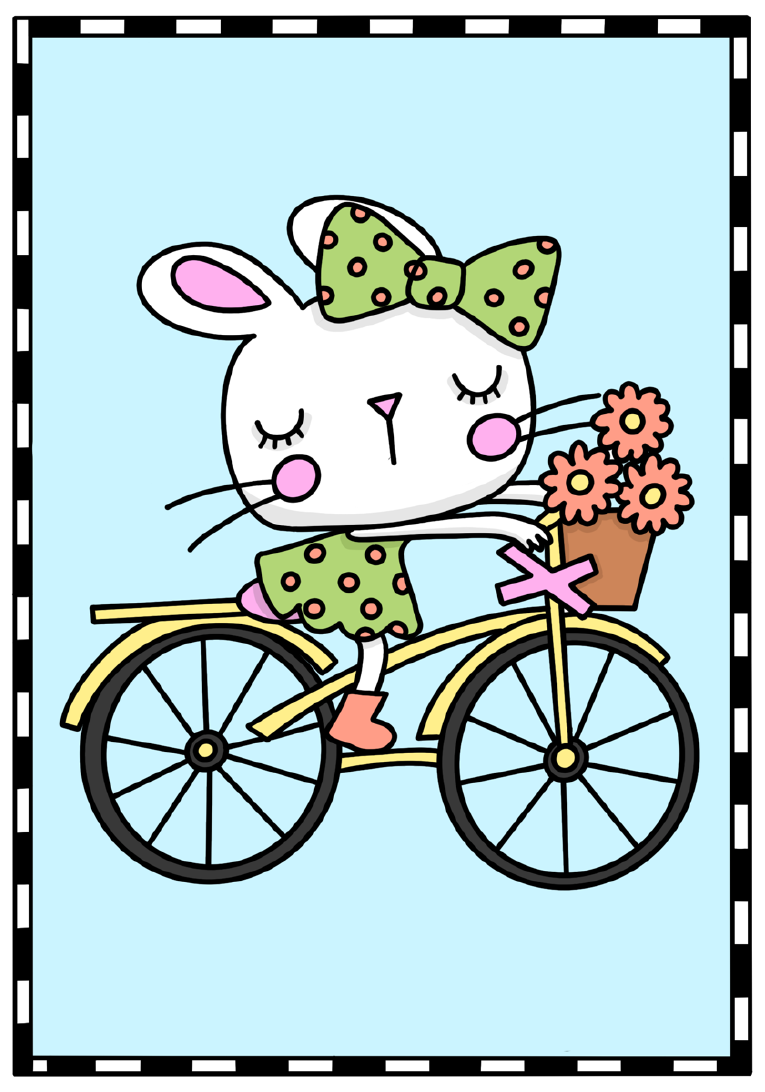}
      {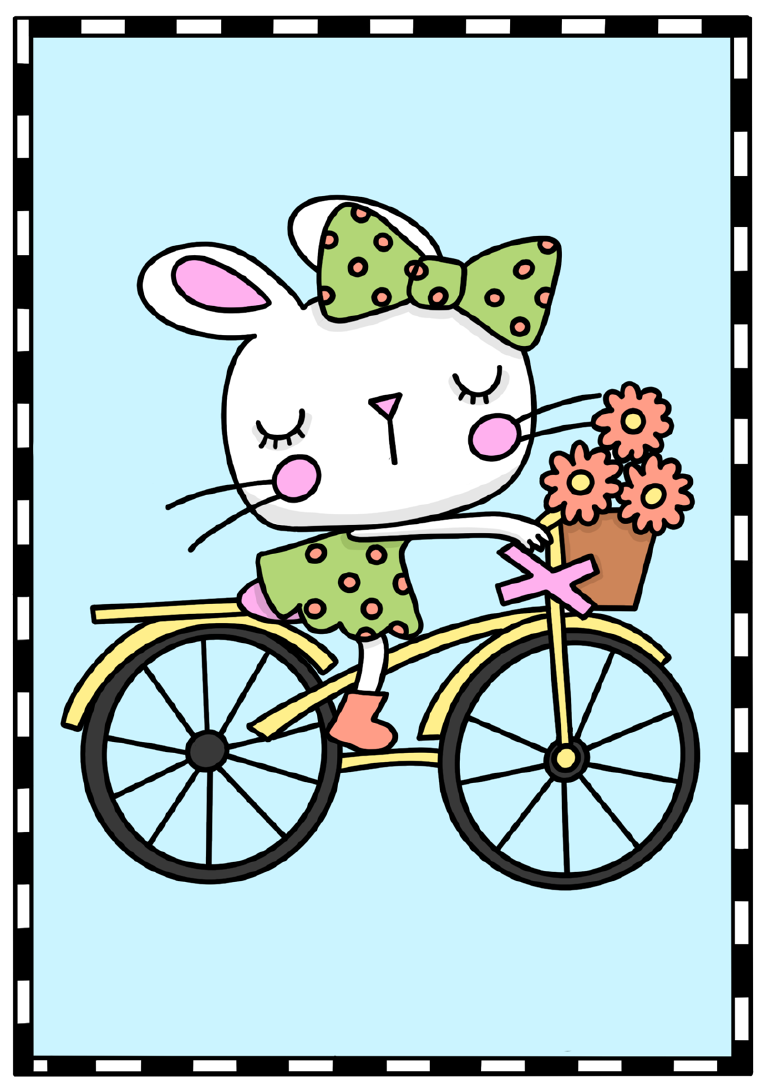}
      {\VDChoice{A}{No difference between A and B.}
       \VDChoice{B}{The ribbon of the rabbit is missing a round pattern in B relative to A.}
       \VDTrue{VDAppear}{C}{The right eye of the rabbit is missing an eyelash in B relative to A.}
       \VDChoice{D}{The bicycle is missing a pedal in B relative to A.}}
      {\VDModels
        {\VDModel{openai.png}{}{\VDNo}}
        {\VDModel{gemini.png}{3.1}{\VDNo}}
        {\VDModel{gemini.png}{3.5}{\VDNo}}
        {\VDModel{kimi.png}{}{\VDNo}}
        {\VDModel{grok.png}{}{\VDYes}}
        {\VDModel{qwen.png}{}{\VDYes}}}
  \end{minipage}\hfill
  \begin{minipage}[t][0.245\textheight][t]{0.492\textwidth}
    \VDPanel{Noise / resolution}{VDNoise}
      {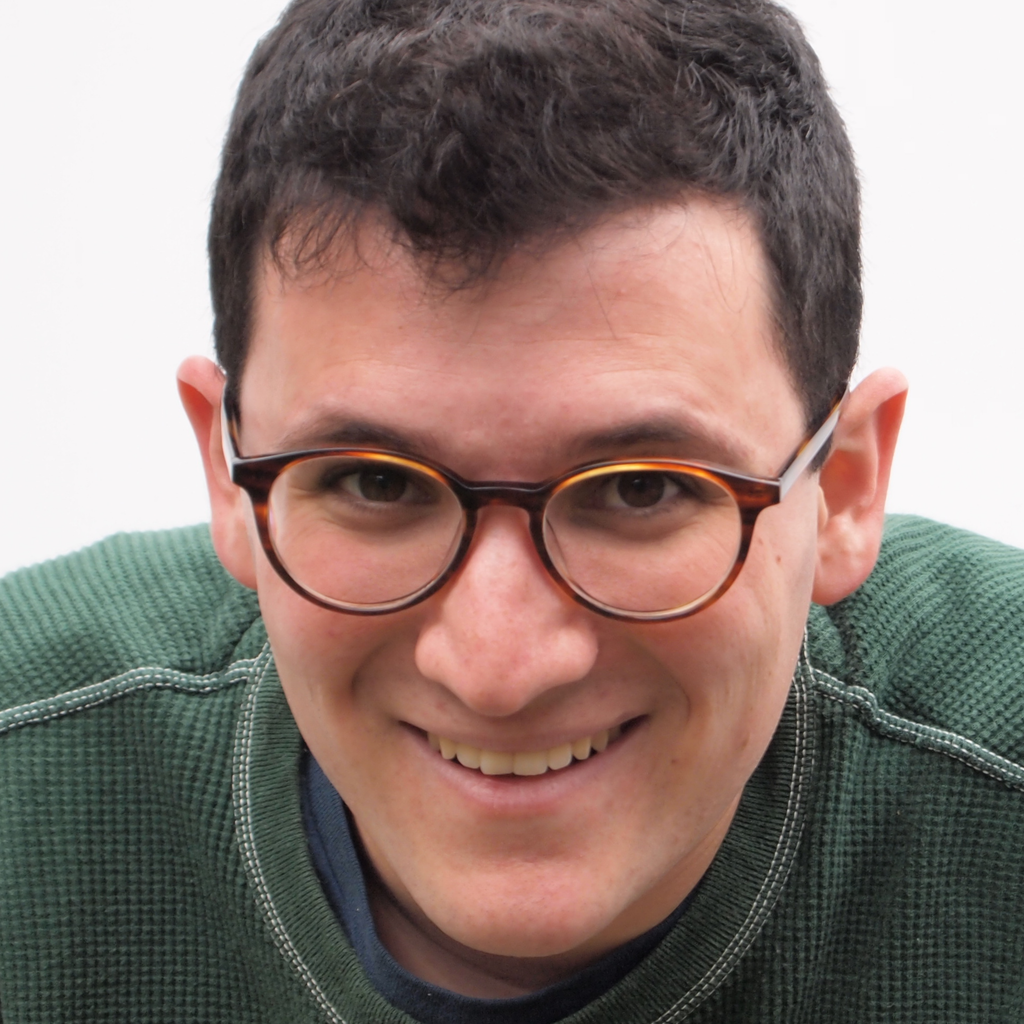}
      {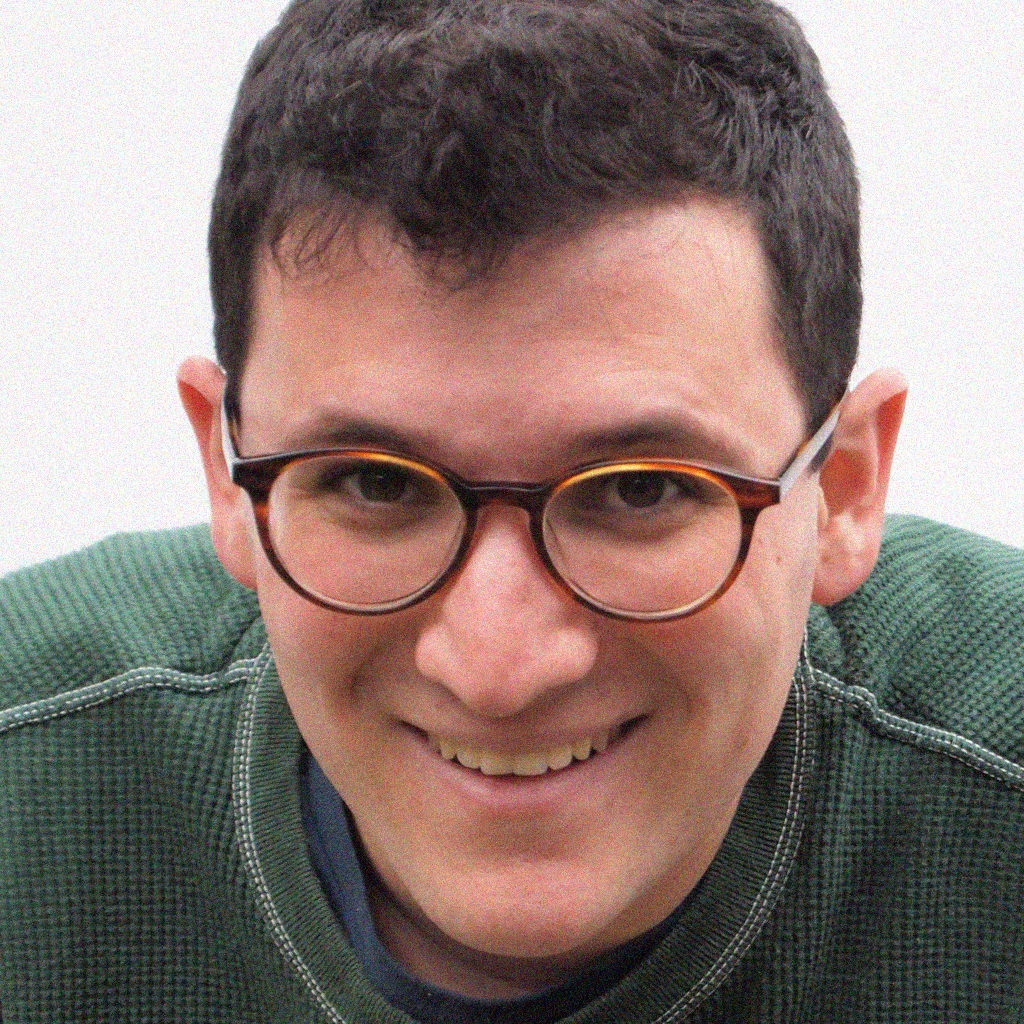}
      {\VDChoice{A}{No difference between A and B.}
       \VDChoice{B}{B has a warmer color temperature than A.}
       \VDChoice{C}{B is clearer than A.}
       \VDTrue{VDNoise}{D}{B is noisier than A.}}
      {\VDModels
        {\VDModel{openai.png}{}{\VDYes}}
        {\VDModel{gemini.png}{3.1}{\VDYes}}
        {\VDModel{gemini.png}{3.5}{\VDYes}}
        {\VDModel{kimi.png}{}{\VDYes}}
        {\VDModel{grok.png}{}{\VDNo}}
        {\VDModel{qwen.png}{}{\VDNo}}}
  \end{minipage}

  \caption{\textbf{Qualitative \textsc{VDiff-Bench} results across all ten change categories (part 1 of 2).} Each panel shows the complete image pair at a common height, all four answer choices with the ground truth colored, and correctness for six representative MLLMs. Green checks and red crosses indicate correct and incorrect selections.}
  \label{fig:qualitative-atlas}
\end{figure}

\renewcommand{\VDImageHeight}{1.55in}
\begin{figure}[!t]
  \ContinuedFloat
  \centering
  \VDLegend
  \vspace{0.8ex}

  \begin{minipage}[t][0.355\textheight][t]{0.492\textwidth}
    \VDPanel{Texture}{VDTexture}
      {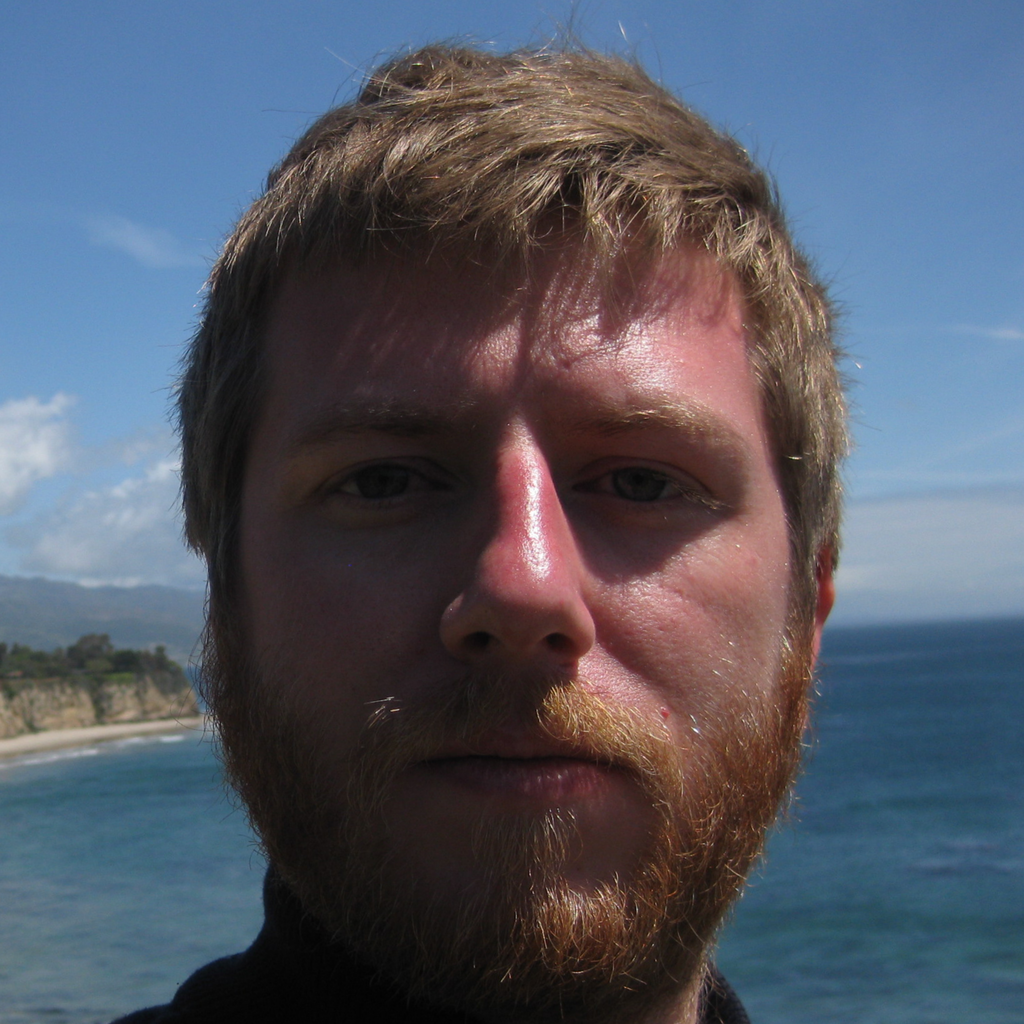}
      {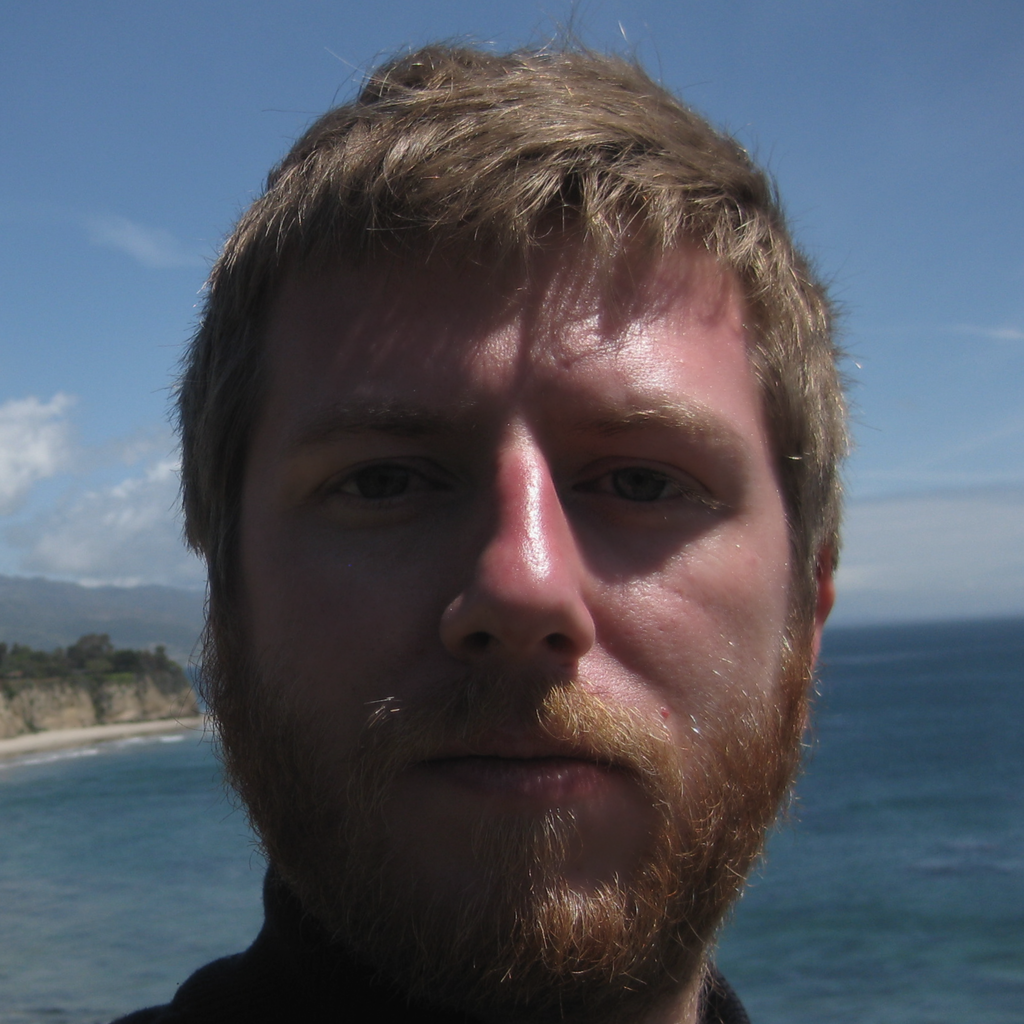}
      {\VDChoice{A}{No difference between A and B.}
       \VDChoice{B}{The man's brows changed to a lighter color in B.}
       \VDTrue{VDTexture}{C}{B is smoother than A.}
       \VDChoice{D}{B is sharper than A.}}
      {\VDModels
        {\VDModel{openai.png}{}{\VDYes}}
        {\VDModel{gemini.png}{3.1}{\VDYes}}
        {\VDModel{gemini.png}{3.5}{\VDYes}}
        {\VDModel{kimi.png}{}{\VDNo}}
        {\VDModel{grok.png}{}{\VDNo}}
        {\VDModel{qwen.png}{}{\VDNo}}}
  \end{minipage}\hfill
  \begin{minipage}[t][0.355\textheight][t]{0.492\textwidth}
    \VDPanel{Substitution / size}{VDSubstitution}
      {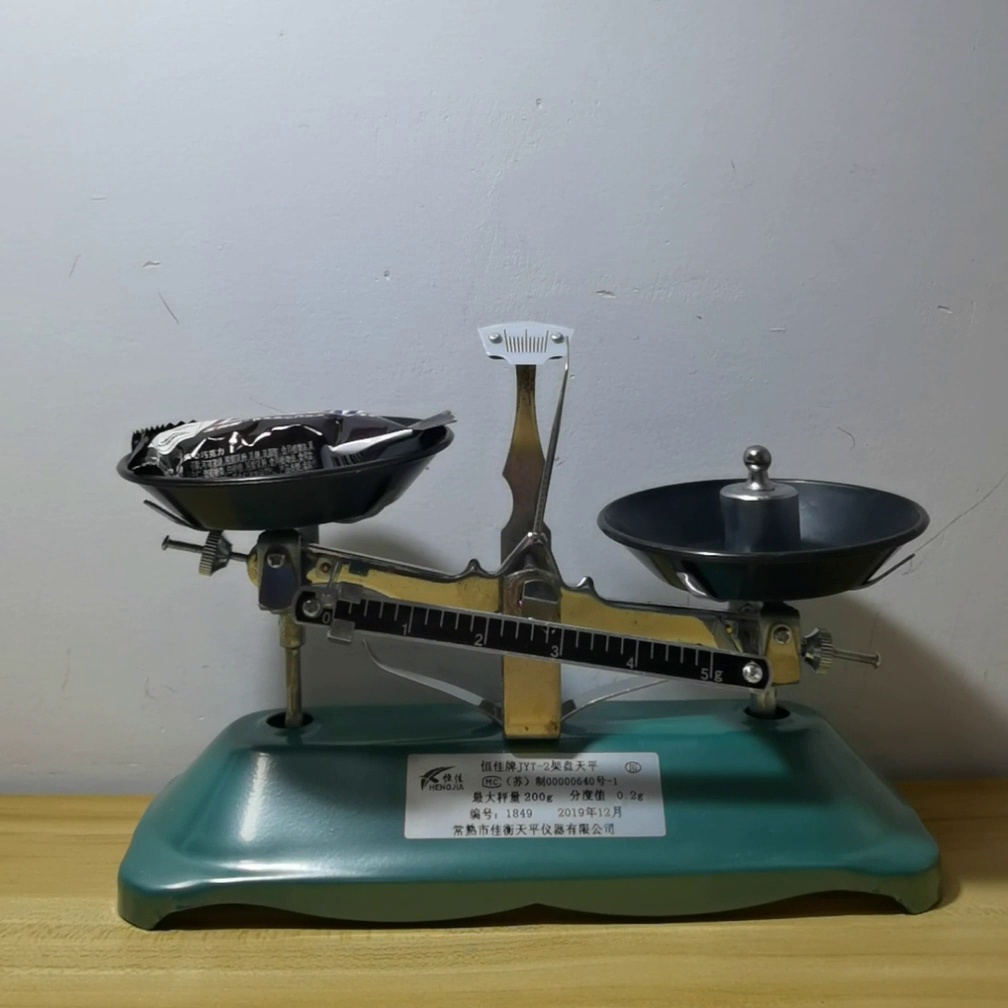}
      {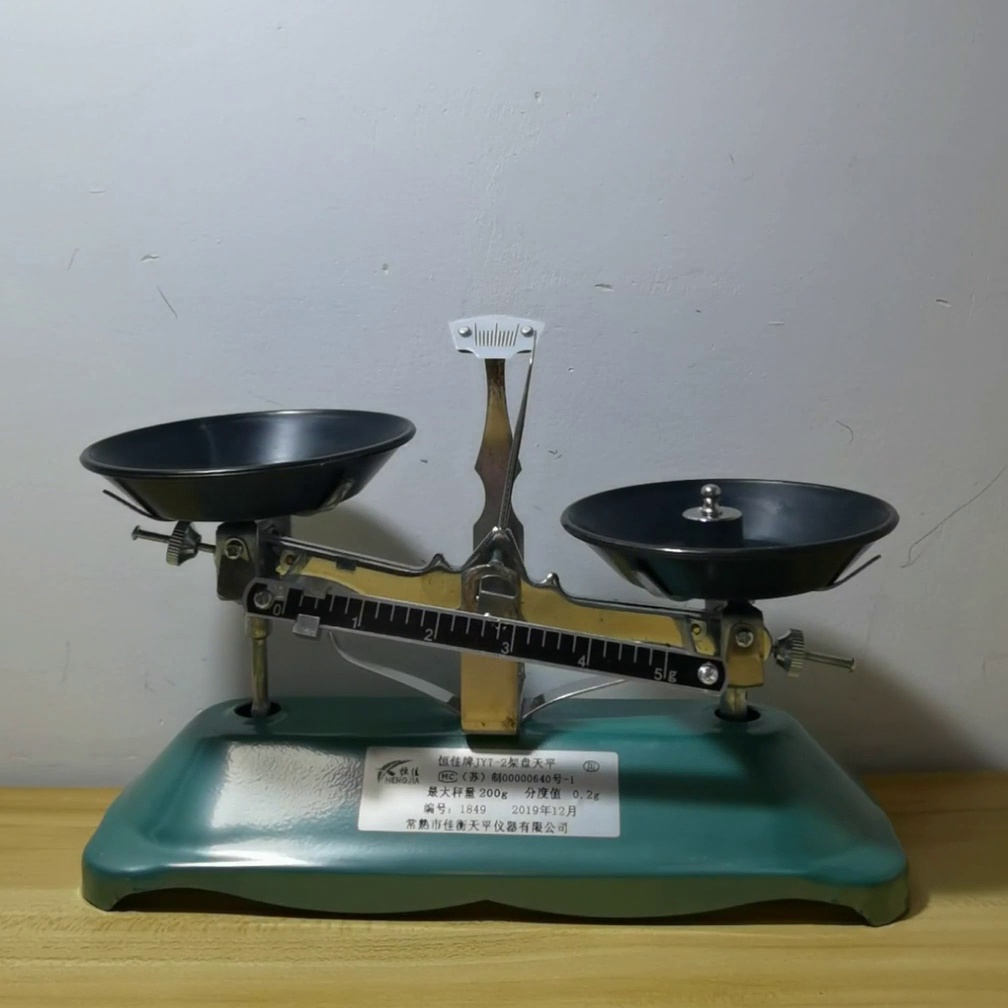}
      {\VDTrue{VDSubstitution}{A}{The large weight on the right tray of the scale in A is changed to a small weight in B.}
       \VDChoice{B}{The large weight on the right tray of the scale in A is changed to a black package in B.}
       \VDChoice{C}{The black package on the left tray of the scale in A is changed to a small weight in B.}
       \VDChoice{D}{No difference between A and B.}}
      {\VDModels
        {\VDModel{openai.png}{}{\VDNo}}
        {\VDModel{gemini.png}{3.1}{\VDYes}}
        {\VDModel{gemini.png}{3.5}{\VDYes}}
        {\VDModel{kimi.png}{}{\VDYes}}
        {\VDModel{grok.png}{}{\VDNo}}
        {\VDModel{qwen.png}{}{\VDNo}}}
  \end{minipage}

  \vspace{1.0ex}
  \begin{minipage}[t][0.355\textheight][t]{0.492\textwidth}
    \VDPanel{OCR / text}{VDOCR}
      {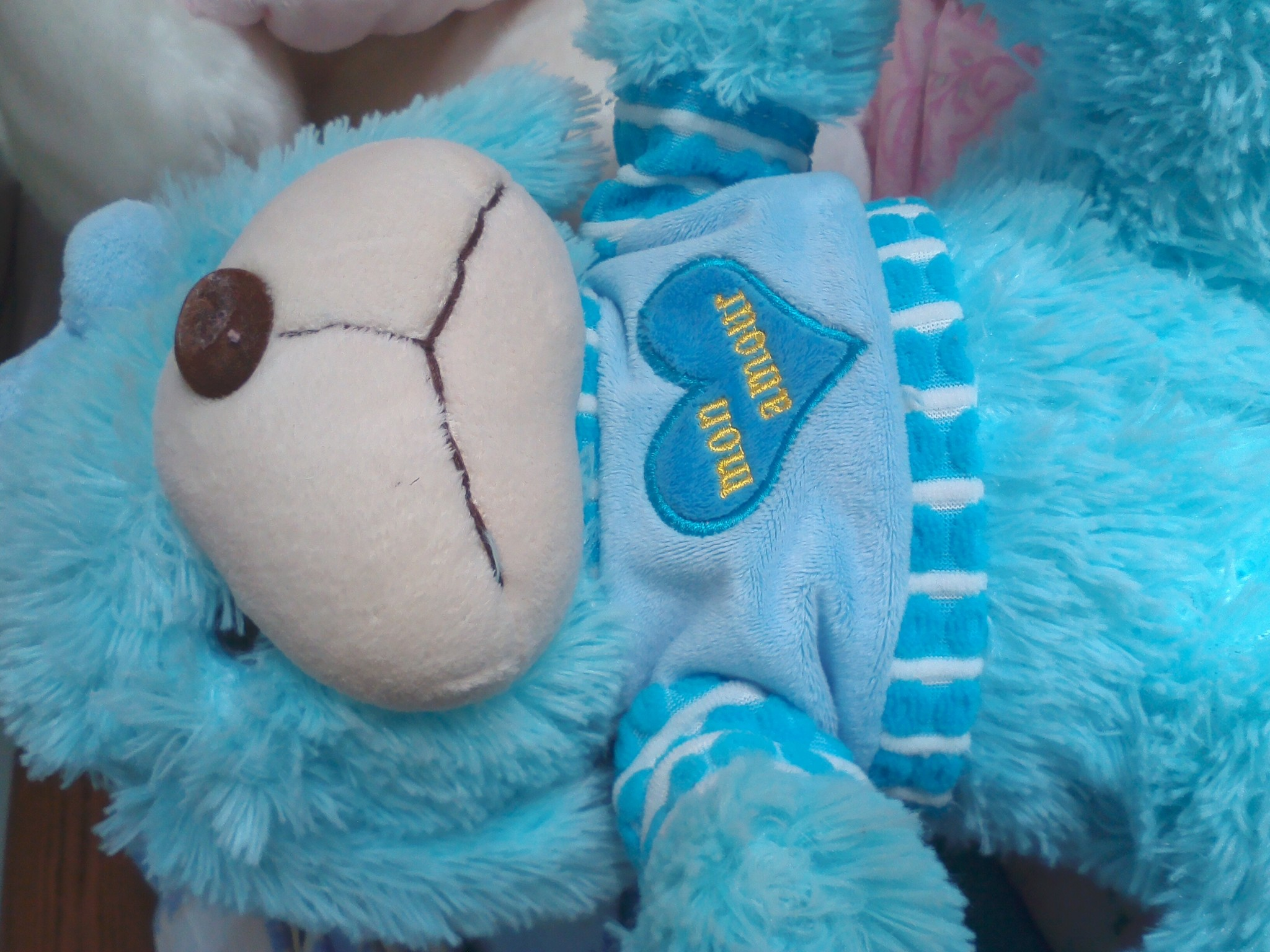}
      {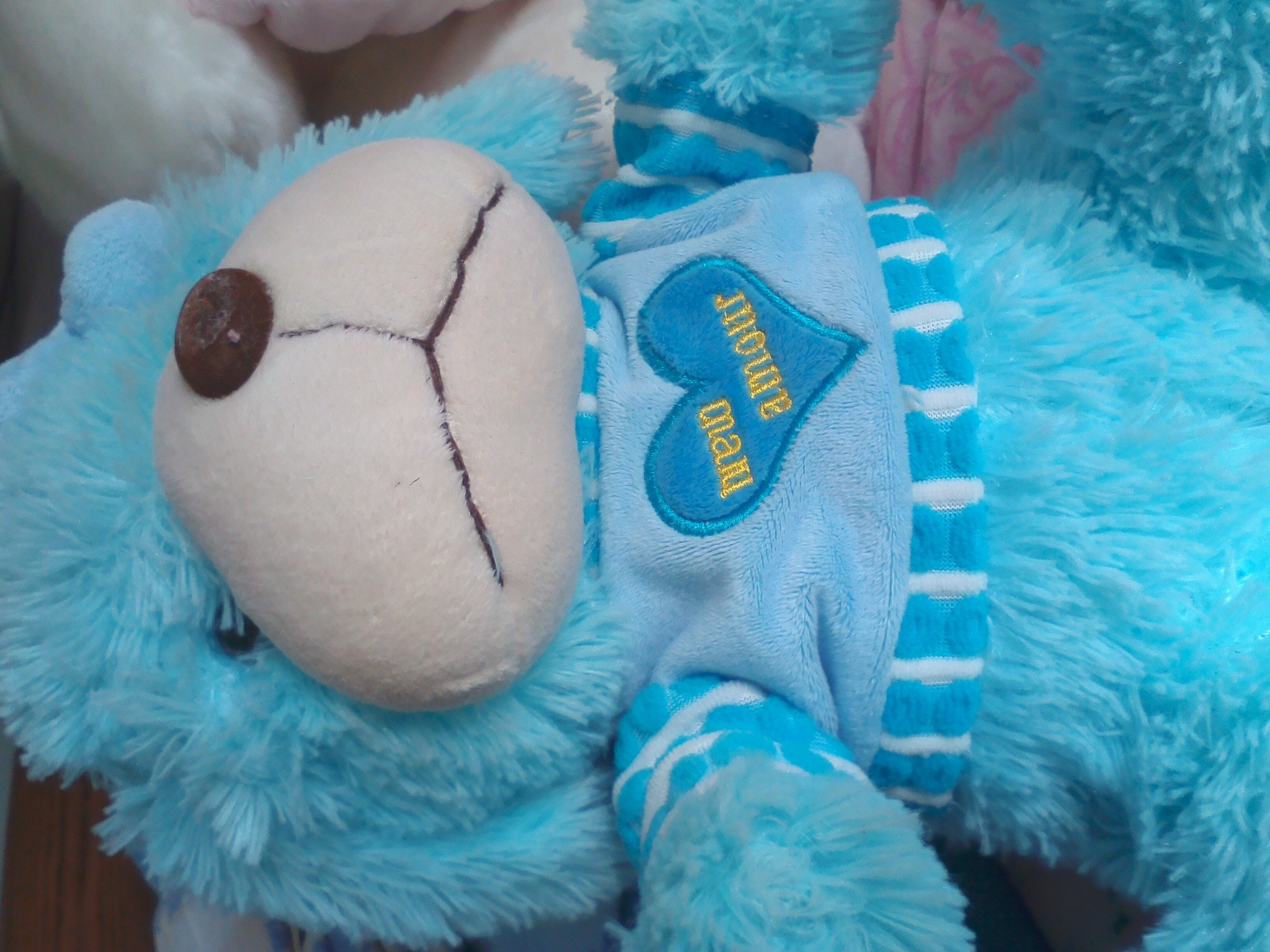}
      {\VDChoice{A}{No difference between A and B.}
       \VDChoice{B}{The Latin text ``amour'' is changed to ``amour'' with an extra letter.}
       \VDChoice{C}{The Latin text ``mon'' is changed to ``men''.}
       \VDTrue{VDOCR}{D}{The Latin text ``mon'' is changed to ``man''.}}
      {\VDModels
        {\VDModel{openai.png}{}{\VDYes}}
        {\VDModel{gemini.png}{3.1}{\VDYes}}
        {\VDModel{gemini.png}{3.5}{\VDNo}}
        {\VDModel{kimi.png}{}{\VDNo}}
        {\VDModel{grok.png}{}{\VDNo}}
        {\VDModel{qwen.png}{}{\VDNo}}}
  \end{minipage}\hfill
  \begin{minipage}[t][0.355\textheight][t]{0.492\textwidth}
    \VDPanel{Illumination}{VDIllumination}
      {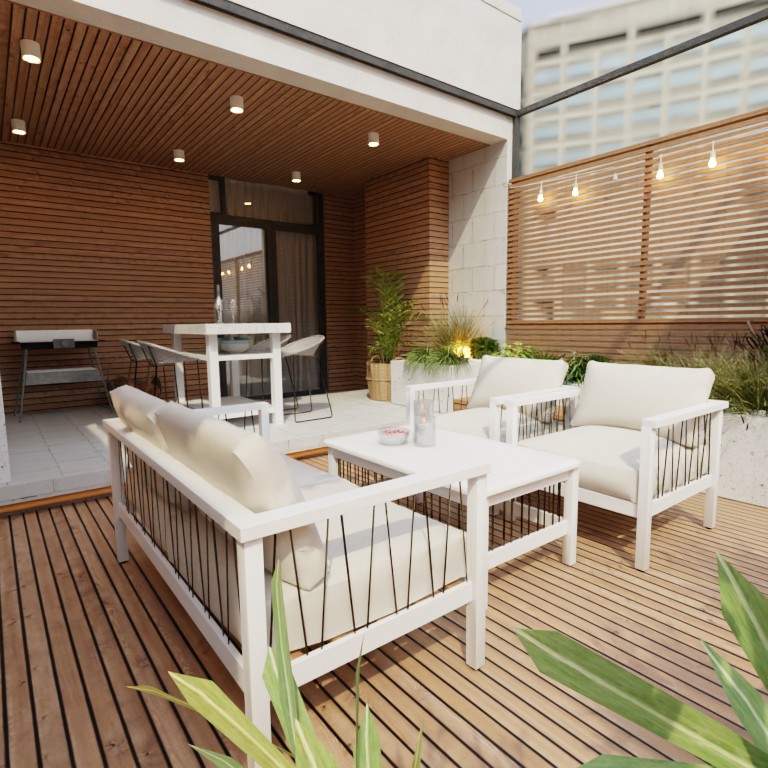}
      {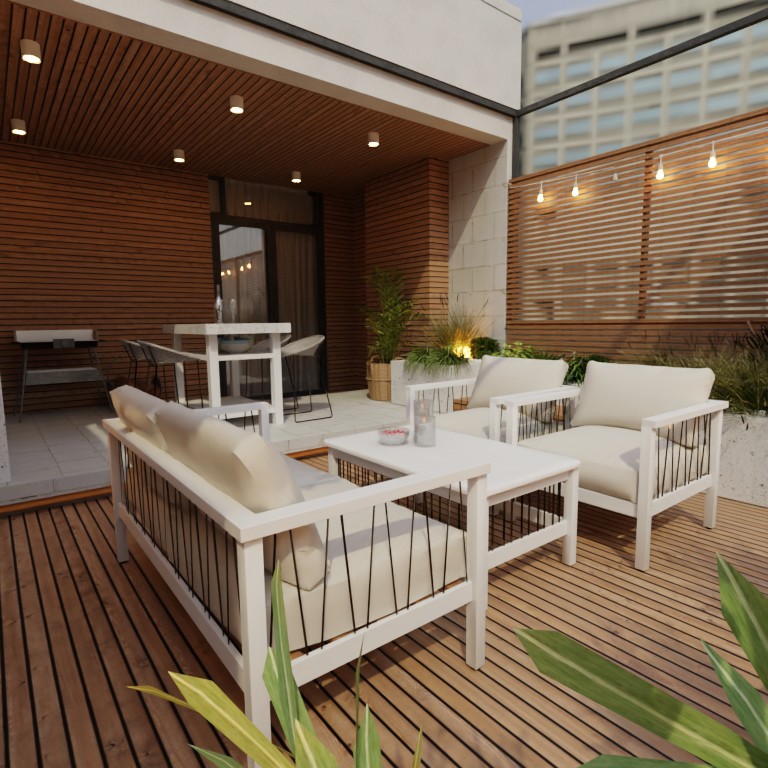}
      {\VDChoice{A}{B is brighter than A.}
       \VDChoice{B}{B is sharper than A.}
       \VDChoice{C}{No difference between A and B.}
       \VDTrue{VDIllumination}{D}{B is darker than A.}}
      {\VDModels
        {\VDModel{openai.png}{}{\VDNo}}
        {\VDModel{gemini.png}{3.1}{\VDYes}}
        {\VDModel{gemini.png}{3.5}{\VDYes}}
        {\VDModel{kimi.png}{}{\VDYes}}
        {\VDModel{grok.png}{}{\VDNo}}
        {\VDModel{qwen.png}{}{\VDNo}}}
  \end{minipage}

  \caption{\textbf{Qualitative \textsc{VDiff-Bench} results across all ten change categories (part 2 of 2).} Layout and symbols follow Figure~\ref{fig:qualitative-atlas}. Examples are selected to illustrate mixed model outcomes rather than category-level prevalence.}
\end{figure}
\endgroup

\end{document}